\documentclass{article} 
\usepackage{iclr2026_conference,times}

\usepackage{amsmath,amsfonts,bm}

\def\eqref#1{equation~\ref{#1}}

\def\1{\bm{1}}

\DeclareMathAlphabet{\mathsfit}{\encodingdefault}{\sfdefault}{m}{sl}
\SetMathAlphabet{\mathsfit}{bold}{\encodingdefault}{\sfdefault}{bx}{n}

\usepackage{hyperref}
\usepackage{url}
\usepackage{graphicx}
\usepackage{booktabs}
\usepackage{colortbl}
\usepackage[table]{xcolor}
\usepackage{multirow}
\usepackage{makecell}
\usepackage{array}
\usepackage{adjustbox}
\usepackage{subcaption}
\usepackage{listings}
\usepackage{xcolor}
\usepackage{float}
\usepackage{stmaryrd}

\title{\includegraphics[height=1.2em]{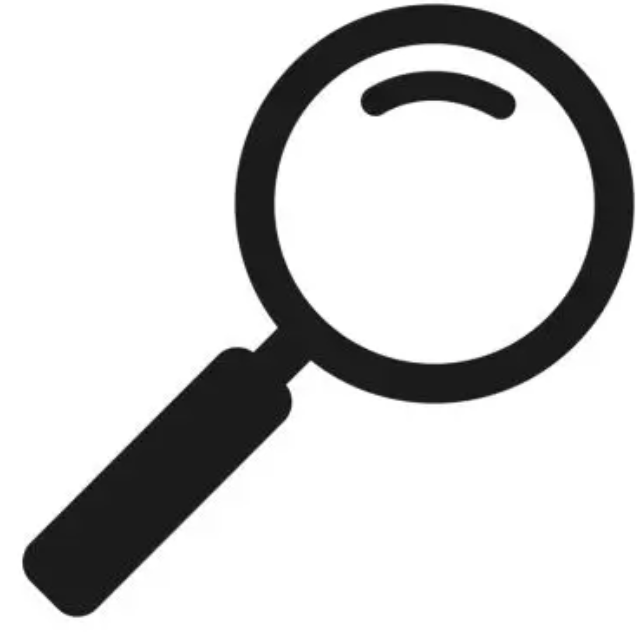} PDE-SHARP: \textbf{PDE} \textbf{S}olver \textbf{H}ybrids through \textbf{A}nalysis and \textbf{R}efinement \textbf{P}asses}

\author{Shaghayegh ~Fazliani \thanks{ Correspondence: \texttt{fazliani@stanford.edu}. 
Code visible at: \url{https://github.com/udellgroup/PDE-SHARP}.} \\
Department of Mathematics\\
Stanford University, Stanford, CA\\
\And
Madeleine Udell \\
Department of Management Science \& Engineering \\
Stanford University, Stanford, CA \\
}

\iclrfinalcopy

\begin{document}

\maketitle

\begin{abstract}
Current LLM-driven approaches using test-time computing to generate PDE solvers execute a large number of solver samples to identify high-accuracy solvers. 
These paradigms are especially costly for complex PDEs requiring substantial computational resources for numerical evaluation.
We introduce PDE-SHARP, a framework to reduce computational costs by replacing expensive scientific computation by cheaper LLM inference that achieves superior solver accuracy with 60-75\% fewer computational evaluations.
PDE-SHARP employs three stages: \textbf{(1) Analysis}: mathematical chain-of-thought analysis including PDE classification, solution type detection, and stability analysis; \textbf{(2) Genesis}: solver generation based on mathematical insights from the previous stage; and \textbf{(3) Synthesis}: collaborative selection-hybridization tournaments in which LLM judges iteratively refine implementations through flexible performance feedback.
To generate high-quality solvers, PDE-SHARP requires fewer than 13 solver evaluations on average compared to 30+ for baseline methods, improving accuracy uniformly across tested PDEs by $4\times$ on average,
and demonstrates robust performance across LLM architectures, from general-purpose to specialized reasoning models. 
\end{abstract}
\vspace{-0.3cm}
\section{Introduction}
Partial Differential Equations (PDEs) are fundamental to scientific modeling across physics, engineering, and computational sciences, yet writing robust numerical solvers requires specialized numerical analysis expertise for PDE-specific implementation and tuning, with limited flexibility as each solver targets specific PDE types. 
The success of deep learning has motivated the development of neural PDE solvers, with Physics-Informed Neural Networks (PINNs)~\citep{raissi2019deep, karniadakis2021physicsinformed} and operator learning methods~\citep{li2020fno} emerging as promising alternatives that leverage neural networks to approximate PDE solutions. However, these approaches require extensive training data, lack interpretability, suffer from generalization limits across PDE families, and offer limited accuracy ~\citep{rahaman2019icml, wang2022when}
The result is an ecosystem of specialized PDE solvers that address particular failure modes without a systematic understanding of underlying limitations~\citep{cuomo2022scientific, krishnapriyan2021characterizingpossiblefailuremodes, zhang2021understanding, wang2021understanding}. 

Meanwhile, large language models (LLMs) have demonstrated remarkable aptitude for complex mathematical and scientific challenges~\citep{romera2024mathematical, tian2024scicode}. Sophisticated code generation frameworks employ Chain-of-Thought (CoT) reasoning~\citep{welleck2024decodingmetagenerationinferencetimealgorithms, wei2023chainofthoughtpromptingelicitsreasoning, kojima2023largelanguagemodelszeroshot}, Mixture-of-Agents (MoA) strategies~\citep{optillm, wang2024mixtureofagentsenhanceslargelanguage}, and advanced inference-time scaling techniques~\citep{snell2024scalingllmtesttimecompute} to achieve state-of-the-art performance across programming tasks. LLM-as-a-judge frameworks~\citep{jiang2025codejudgebenchbenchmarkingllmasajudgecoding, zheng2023judgingllmasajudgemtbenchchatbot} typically employ predetermined evaluation rubrics.
However, PDE solver evaluation presents unique challenges requiring assessment of mathematical correctness, numerical stability, computational efficiency, and domain-specific accuracy, factors that demand context-dependent evaluation criteria rather than static rubrics, as optimal trade-offs and performance standards vary significantly across PDE families and application domains. The task of creating reliable solver codes for PDEs sits at the intersection of applied mathematics, numerical analysis, and code generation, making it an ideal testbed to evaluate LLMs' mathematical and technical capabilities. Current approaches fall into two general categories. 1) Fine-tuning methods specialize models for mathematical reasoning~\citep{lu2024mathcoder2bettermathreasoning} and subsequent domain-specific adaptation to particular PDE families~\citep{soroco2025pdecontrollerllmsautoformalizationreasoning}. These require substantial computational resources for multi-stage training and offer limited generalizability across PDE types. 2) Inference-only frameworks using general-purpose LLMs and techniques such as automated debugging \citep{chen2023teachinglargelanguagemodels}, self-refinement \citep{madaan2023selfrefineiterativerefinementselffeedback}, and test-time scaling \citep{snell2024scalingllmtesttimecompute}.
CodePDE~\citep{li2025codepdeinferenceframeworkllmdriven} avoids fine-tuning but relies on brute-force sampling strategies, generating and executing 30+ solver candidates to identify optimal solutions. This paradigm becomes especially costly for complex PDEs requiring high-performance computing resources for numerical evaluation.

To address these limitations, we introduce \textbf{PDE-SHARP}, an LLM-driven PDE solver generation framework that achieves superior accuracy with 60-75\% fewer computational evaluations ---  through
intelligent generation rather than exhaustive sampling --- in three stages: \textbf{(1) Analysis} analyzes the PDE through structured questions to develop a numerically-stable solver plan; \textbf{(2) Genesis} generates solver candidates without immediate execution; \textbf{(3) Synthesis} uses LLM judges to iteratively select, execute, and refine solvers based on provided performance feedback in each round. With this approach, PDE-SHARP swaps inexpensive LLM inference for expensive scientific computation, only executing refined solvers each round. This exchange is worthwhile for computationally intensive PDEs for which GPU/HPC resources dominate costs.

\begin{figure}[h!]
    \centering
\includegraphics[width=\linewidth]{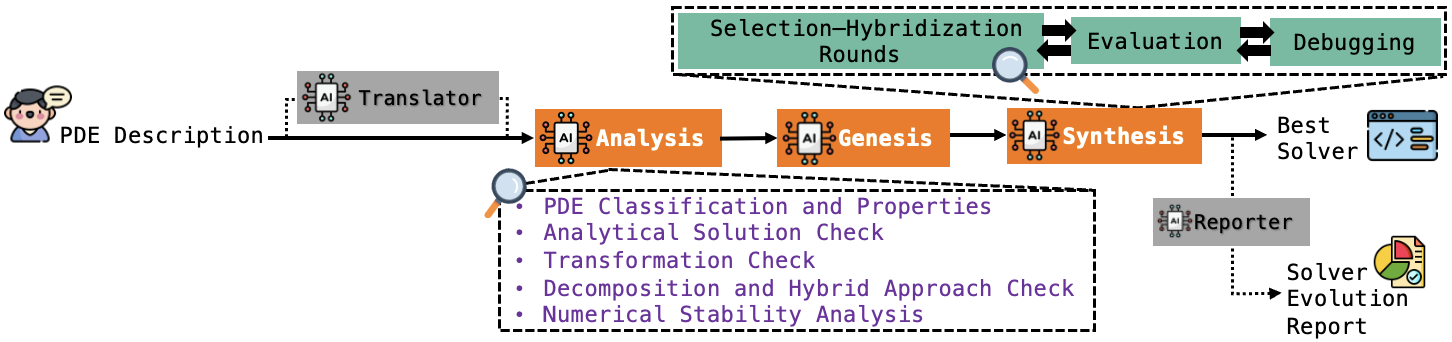}
    \caption{PDE-SHARP framework overview. The three core stages are Analysis, Genesis, and Synthesis. Optional components (Translator, Reporter) enhance usability as explained in section \ref{PDE-SHARP Framework}. PDE-SHARP generates higher accuracy solvers with 60-75\% fewer solver evaluations compared to tested baselines.}
    \label{PDE-SHARP Main Framework}
\end{figure}

\paragraph{Contributions.} The experimental results highlight PDE-SHARP's key contributions:
\begin{itemize}
\item \textbf{Computational Efficiency.} PDE-SHARP reduces expensive solver evaluations by 60-75\% (requiring fewer than 13 solver evaluations on average compared to 30+ in best-of-$n$ baselines) while achieving superior solution accuracy, demonstrating considerable resource savings for complex simulations.
\item \textbf{Mathematical Analysis.} PDE-specific mathematical chain-of-thought reasoning with targeted stability analysis produces mathematically-informed solver strategies, leading to higher initial code quality compared to generic template-based generation.
\item \textbf{Collaborative LLM Tournaments.} PDE-SHARP's synthesis phase improves on standard test-time computing approaches by 4$\times$ on average  using fewer evaluations.
\item \textbf{Enhanced Implementation Quality.} Experiments indicate PDE-SHARP solvers achieve bug-free execution in 63-67\% fewer debugging iterations (0.33 vs. 0.9-1.4 iterations per solver) and enjoy superior numerical convergence properties.
\item \textbf{Robustness to LLM Choice.} PDE-SHARP achieves more consistent performance across diverse LLM types (general-purpose, coding-specific, reasoning models) compared to the baselines, showing robustness to the underlying code generator LLM choice.
\item \textbf{Flexible Feedback Integration.} PDE-SHARP can improve solvers using several feedback mechanisms --- solution-based metrics (relative error), physics-based metrics (PDE residual),
and no feedback --- to adapt to research scenarios from benchmark validation with known solutions to real-world cases with limited simulation data or physics-only assessments.
\end{itemize}
\vspace{-0.3cm}
\section{Background \& Related Work} \label{sec:related}
\vspace{-0.2cm}
\textbf{Classical Solvers \& Neural Methods.} Traditional numerical methods for PDE solving, e.g. finite difference, finite element, and spectral methods, require considerable domain expertise for effective implementation ~\citep{Strang:2007, leveque2007finite}. Modern scientific computing frameworks such as FEniCS ~\citep{Alnaes2015}, deal.II~\citep{Arndt_2021} for finite element, and PETSc~\citep{petsc-web-page} have facilitated access to these methods for broad PDE classes. However, 1) considerable numerical analysis knowledge is still required for optimal performance; and 2) general approaches fail at exploiting PDE-specific mathematical structure to achieve superior performance. The key challenge is thus identifying which approach suits a particular PDE without extensive domain expertise. 

The success of deep learning has motivated extensive research into neural PDE solvers. PINNs variants~\citep{raissi2019deep, wang2022when} approximate PDE solutions through residual minimization. Physics-informed operator learning methods~\citep{li2020fno, lu2021deeponet} learn solution operators rather than individual solutions, offering improved generalization.
Feature engineering techniques such as random Fourier features~\citep{wang2021eigenvector, fazliani2025enhancingphysicsinformedneuralnetworks}, residual-based attention~\citep{anagnostopoulos2023residualbasedattentionconnectioninformation}, and radial basis functions~\citep{zeng2024featuremappingphysicsinformedneural} have further enhanced neural solver capabilities. Foundation models leverage transformer architectures for multiphysics problems~\citep{mccabe2024multiplephysicspretrainingphysical, hao2024dpotautoregressivedenoisingoperator, shen2024upsefficientlybuildingfoundation, herde2024poseidonefficientfoundationmodels}. These neural approaches, however, require extensive training data, lack transparency and interpretability regarding solution generation processes, and have generalization limits.

Custom solver generation offers several advantages over neural surrogates and black-box library usage: full algorithmic transparency enables targeted PDE-specific optimization, simplified debugging and modification, and direct control over every detail. This is crucial when solver behavior needs explanation or when problem-specific modifications are required.

\textbf{LLM-Driven Code Generation for PDEs.} The integration of LLMs into scientific computing has emerged along two primary paradigms. First is fine-tuning models pretrained on mathematical tasks for domain-specific applications. MathCoder2~\citep{lu2024mathcoder2bettermathreasoning} demonstrates improved mathematical reasoning through continued training. PDE-Controller~\citep{soroco2025pdecontrollerllmsautoformalizationreasoning} continues this approach by fine-tuning MathCoder2-DeepSeekMath on specific PDE families such as heat and wave equations. While effective for targeted applications, this paradigm requires substantial computational resources for multi-stage training and limits generalizability across diverse PDE types. Second is leveraging inference-time optimization techniques to enhance performance. CodePDE~\citep{li2025codepdeinferenceframeworkllmdriven} implements automated debugging and test-time sampling for diverse solver generation. Frameworks such as OptiLLM~\citep{optillm} integrate multiple inference optimization strategies including Chain-of-Thought (CoT), Mixture-of-Agents (MoA), self-reflection, PlanSearch, etc. These approaches typically rely on computationally expensive best-of-$n$ sampling strategies, generating and evaluating large numbers of solver candidates to identify optimal solutions, which becomes prohibitive for complex PDEs requiring substantial evaluation resources.

Both paradigms face fundamental limitations in balancing solution quality with computational efficiency, motivating the need for more intelligent synthesis approaches that leverage mathematical reasoning without exhaustive sampling or extensive fine-tuning requirements.
\vspace{-0.1cm}
\section{PDE-SHARP Framework} \label{PDE-SHARP Framework}
\begin{figure}[h!]
    \centering
\includegraphics[width=\linewidth]{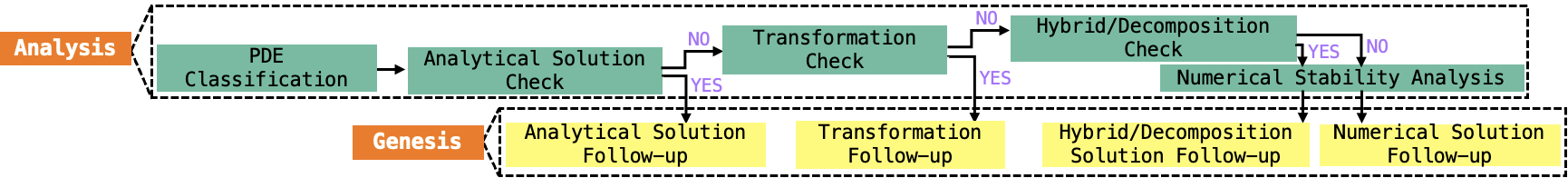}
    \caption{PDE-SHARP Analysis and Genesis stages.}
    \label{PDE-SHARP Analysis Genesis}
\end{figure}

\vspace{-0.05cm}
\paragraph{Stage 1: Analysis.} PDE-SHARP conducts a systematic five-step mathematical analysis to guide solver generation. The process begins with PDE classification (order, linearity, type, boundary conditions) that informs all subsequent decisions. Sequential checks determine if analytical solutions exist, whether transformations can simplify the problem, and if operator decomposition (e.g., separating diffusion and reaction terms) is viable. Each step either directs the framework toward specialized solution strategies in Stage 2 or continues to the next analysis step as shown in Figure \ref{PDE-SHARP Analysis Genesis}. The final stability analysis computes symbolic time-step bounds and selects numerically stable schemes, performed before hybrid/numerical solver generation to ensure robustness. Ablation studies (Appendix \ref{PDE-SHARP Ablation Studies}) demonstrate the effectiveness of this multi-step paradigm over other alternatives.

\paragraph{Stage 2: Genesis.}  PDE solver code is generated using information from the Analysis stage.
\vspace{-0.35cm}
\paragraph{Stage 3: Synthesis} This stage uses Selection–Hybridization Tournaments with LLM judges to iteratively refine solver implementations. Numerical accuracy of the solver can inform judge decisions through a configurable feedback mechanism. 
Synthesis consists of two main steps:

\begin{figure}[h!]
    \centering
\includegraphics[width=\linewidth]{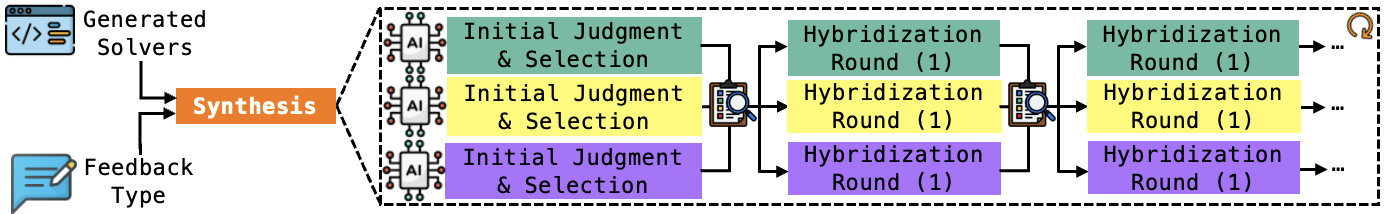}
    \caption{PDE-SHARP Synthesis. This stage can be repeated to address performance saturation.}
    \label{PDE-SHARP Synthesis}
\end{figure}
 
\textbf{(i) Initial Judgment \& Selection:} Given the $n$ generated initial solvers and a specified feedback type, each judge LLM produces a selection of its top $\frac{n}{2}$ choices from the initial list with reasoning behind each choice (prompt format detailed in Appendix \ref{Stage 3: Synthesis}). Each judge also designates one solver from its top $\frac{n}{2}$ list as a nominee for execution and evaluation using the allowed feedback. 

\textbf{(ii) Hybridization Rounds:} The three nominated base solvers are executed and their performance results are shared with all judges. Each judge then proposes modifications to their base solver using a diff/patch format to ensure incremental changes that preserve working code structure and encourage local fixes, with technical justification for each modification. The modified solvers are executed and results again shared with all judges. This process repeats until performance improvements saturate across consecutive rounds or as specified by the user.

When performance improvements saturate or the maximum number of hybridization rounds is reached, the framework initiates another \textbf{judging cycle} that repeats steps (i) and (ii) with an expanded solver set including all previously generated hybrids, their technical justifications, and performance feedback from previous rounds. Judges maintain context within each cycle but reset between cycles, evaluating the expanded set from scratch, to encourage exploration of new strategies.

\textbf{Feedback Types.} The Synthesis stage can incorporate different performance metrics to guide judge decisions during tournaments. We discuss three feedback types: (1) nRMSE: normalized root mean squared error against reference solutions; (2) {PDE residual feedback:} physics-based residual computation that requires no reference data; and (3) {no feedback:} judges rely purely on code analysis. The choice of feedback type allows adaptation to different research scenarios --- from benchmark validation with known solutions to real-world cases with limited reference data. PDE-specific feedback types and their combinations could also be employed for domain-specific optimization. Additional discussions and results appear in Appendix~\ref{Hybridization Feedback Type}. 

\textbf{Optional Stages.} PDE-SHARP includes two optional components for enhanced usability (Figure \ref{PDE-SHARP Main Framework}): \textbf{Translator} converts natural language PDE descriptions into the structured mathematical templates required by the Analysis stage. When user input lacks necessary detail, it requests additional information before proceeding. Users can alternatively bypass this stage by directly providing pre-formatted templates. \textbf{Reporter} generates detailed reports on solver evolution throughout the tournament process, enhancing framework interpretability. An example of such report for the reaction-diffusion PDE is provided in Appendix \ref{Example of a Synthesis Stage Tournament Report: Reaction-Diffusion PDE Solver Refinement}. These reports can serve as feedback for subsequent runs on the same problem, enabling iterative refinement strategies.
\vspace{-0.3cm}
\section{Experiments}\label{Experiments}
We compare PDE-SHARP against multiple baseline methods across five representative PDE tasks from PDEBench ~\citep{takamoto2024pdebenchextensivebenchmarkscientific} (Table~\ref{pdes}). Discussions on neural methods and some LLM-driven approaches (agentic workflows, fine-tuned mathematical models, etc.) appear in Appendix \ref{Additional Experimental Setups}. In our experiments, we focus on LLM-driven  baselines using test-time computing for code generation that directly compete with PDE-SHARP's approach. \textbf{CodePDE} ~\citep{li2025codepdeinferenceframeworkllmdriven} generates solvers using chain-of-thought prompting and executes all samples to report the best performance. A refined variant, \textbf{CodePDE-R}, is also tested as a baseline. \textbf{OptiLLM} ~\citep{optillm} implements inference optimization techniques including Chain-of-Thought (CoT), Mixture-of-Agents (MoA), and Cerebras Planning and Optimization (CePO). Experimental details appear in Appendix \ref{Additional Experimental Setups}.

\begin{table}[htbp]
  \centering
  \caption{Tested PDEs; details in Appendix~\ref{Additional Details on the Tested PDEs}.  \textbf{Dimension} column indicates the \textit{spatial dimension} and NL stands for non-linear in the table.}
  \vspace{0.2cm}
  \label{pdes}
  \begin{tabular}{l c c c c}
    \toprule
    \textbf{PDE} & \textbf{Dimension} & \textbf{Type} & \textbf{State} & \textbf{Solution Behavior} \\
    \midrule
    Advection         & 1D & Linear     & Time‑dependent & Smooth \\
    Burgers           & 1D & Highly NL  & Time‑dependent & Shock‑forming \\
    Reaction‑Diffusion & 1D & Mildly NL & Time‑dependent & Smooth \\
    Navier‑Stokes     & 1D & Highly NL  & Time‑dependent & Shock‑forming \\
    Darcy Flow        & 2D & Mildly NL  & Steady‑state   & Smooth \\
    \bottomrule
  \end{tabular}
\end{table}


\textbf{Experimental Setup:} All methods generate $n = 32$ initial solver candidates for fair comparison (Appendix \ref{Test-time Scaling for PDE-SHARP}). Baselines execute all candidates (CodePDE-R executes 44 with refinements). PDE-SHARP uses three judge LLMs (Appendix \ref{Structure of the Tournaments}) in collaborative tournaments, executing only refined candidates per hybridization round. For Section \ref{Experiments} experiments, PDE-SHARP uses nRMSE on 100 validation samples as tournament feedback. All methods are evaluated on a separate test set of 100 random PDEBench samples per PDE task (Table \ref{tab:nrmse-results}). Additional feedback types and judge configurations appear in Appendix \ref{PDE-SHARP Ablation Studies}.


\subsection{Results \& Analysis} \label{Results and Analysis}

 Table \ref{tab:nrmse-results} shows solver accuracy across all PDEs and baselines. 

 \begin{table}[H]
  \centering
  \caption{PDE-SHARP improves solver accuracy and is robust to choice of LLM. Solution accuracy is measured by nRMSE relative to the reference solution from PDEBench. Cell colors 
  use a colormap log-normalized independently within each PDE column to highlight per-task variation. }
  \label{tab:nrmse-results}
  \resizebox{\textwidth}{!}{%
\begin{tabular}{llccccc}
\toprule
 & & Advection & Burgers & Reaction-Diffusion & Navier-Stokes & Darcy \\
\midrule
\multirow{6}{*}{\textbf{OptiLLM-CoT}}
 & Gemma 3 & \cellcolor[RGB]{240,131,91} 5.34e-03 & \cellcolor[RGB]{236,102,92} 5.32e-02 & \cellcolor[RGB]{244,156,90} 2.07e-01 & \cellcolor[RGB]{246,174,90} 9.58e-02 & \cellcolor[RGB]{248,200,125} 8.01e-02 \\
 & LLaMA 3.3 & \cellcolor[RGB]{236,102,92} 7.71e-03 & \cellcolor[RGB]{237,109,92} 4.38e-02 & \cellcolor[RGB]{243,153,90} 2.24e-01 & \cellcolor[RGB]{237,108,92} 2.42e-01 & \cellcolor[RGB]{236,102,92} 1.01e+00 \\
 & Qwen 3 & \cellcolor[RGB]{241,142,90} 4.67e-03 & \cellcolor[RGB]{250,231,168} 1.52e-03 & \cellcolor[RGB]{236,102,92} 9.38e-01 & \cellcolor[RGB]{236,102,92} 2.63e-01 & \cellcolor[RGB]{238,121,91} 6.34e-01 \\
 & DeepSeek-R1 & \cellcolor[RGB]{241,137,91} 4.97e-03 & \cellcolor[RGB]{175,216,170} 3.04e-04 & \cellcolor[RGB]{243,150,90} 2.45e-01 & \cellcolor[RGB]{247,184,104} 8.34e-02 & \cellcolor[RGB]{150,207,164} 5.34e-03 \\
 & GPT-4o & \cellcolor[RGB]{249,221,154} 1.72e-03 & \cellcolor[RGB]{249,220,152} 2.12e-03 & \cellcolor[RGB]{250,235,173} 2.23e-02 & \cellcolor[RGB]{238,121,91} 2.01e-01 & \cellcolor[RGB]{237,109,92} 8.51e-01 \\
 & o3 & \cellcolor[RGB]{214,231,178} 9.74e-04 & \cellcolor[RGB]{193,223,174} 4.08e-04 & \cellcolor[RGB]{243,154,90} 2.21e-01 & \cellcolor[RGB]{235,239,183} 3.12e-02 & \cellcolor[RGB]{151,207,165} 5.47e-03 \\
\midrule
\multirow{6}{*}{\textbf{OptiLLM-MoA}}
 & Gemma 3 & \cellcolor[RGB]{243,155,90} 3.97e-03 & \cellcolor[RGB]{247,195,117} 4.21e-03 & \cellcolor[RGB]{244,163,89} 1.74e-01 & \cellcolor[RGB]{248,199,123} 6.78e-02 & \cellcolor[RGB]{249,221,154} 4.69e-02 \\
 & LLaMA 3.3 & \cellcolor[RGB]{245,243,185} 1.23e-03 & \cellcolor[RGB]{247,190,112} 4.71e-03 & \cellcolor[RGB]{245,168,89} 1.49e-01 & \cellcolor[RGB]{237,111,92} 2.29e-01 & \cellcolor[RGB]{244,163,89} 2.13e-01 \\
 & Qwen 3 & \cellcolor[RGB]{218,233,179} 1.01e-03 & \cellcolor[RGB]{182,219,171} 3.45e-04 & \cellcolor[RGB]{247,184,103} 9.68e-02 & \cellcolor[RGB]{168,214,168} 1.79e-02 & \cellcolor[RGB]{147,206,164} 5.12e-03 \\
 & DeepSeek-R1 & \cellcolor[RGB]{214,231,178} 9.74e-04 & \cellcolor[RGB]{162,212,167} 2.49e-04 & \cellcolor[RGB]{245,169,89} 1.48e-01 & \cellcolor[RGB]{158,210,166} 1.65e-02 & \cellcolor[RGB]{145,205,164} 5.01e-03 \\
 & GPT-4o & \cellcolor[RGB]{248,209,137} 2.01e-03 & \cellcolor[RGB]{161,211,167} 2.41e-04 & \cellcolor[RGB]{250,241,180} 1.94e-02 & \cellcolor[RGB]{211,230,177} 2.56e-02 & \cellcolor[RGB]{145,205,164} 5.02e-03 \\
 & o3 & \cellcolor[RGB]{249,220,153} 1.74e-03 & \cellcolor[RGB]{172,215,169} 2.91e-04 & \cellcolor[RGB]{244,156,90} 2.09e-01 & \cellcolor[RGB]{138,203,162} 1.39e-02 & \cellcolor[RGB]{147,206,164} 5.07e-03 \\
\midrule
\multirow{6}{*}{\textbf{OptiLLM-CePO}}
 & Gemma 3 & \cellcolor[RGB]{244,159,90} 3.74e-03 & \cellcolor[RGB]{247,196,120} 4.01e-03 & \cellcolor[RGB]{244,159,90} 1.89e-01 & \cellcolor[RGB]{248,204,130} 6.32e-02 & \cellcolor[RGB]{249,226,161} 4.12e-02 \\
 & LLaMA 3.3 & \cellcolor[RGB]{231,237,182} 1.11e-03 & \cellcolor[RGB]{247,192,114} 4.53e-03 & \cellcolor[RGB]{246,171,89} 1.36e-01 & \cellcolor[RGB]{238,116,92} 2.18e-01 & \cellcolor[RGB]{245,165,89} 1.98e-01 \\
 & Qwen 3 & \cellcolor[RGB]{218,233,179} 1.01e-03 & \cellcolor[RGB]{178,217,170} 3.23e-04 & \cellcolor[RGB]{247,186,106} 8.91e-02 & \cellcolor[RGB]{179,218,171} 1.97e-02 & \cellcolor[RGB]{144,205,163} 4.83e-03 \\
 & DeepSeek-R1 & \cellcolor[RGB]{213,230,178} 9.71e-04 & \cellcolor[RGB]{161,211,167} 2.43e-04 & \cellcolor[RGB]{246,170,89} 1.39e-01 & \cellcolor[RGB]{168,214,168} 1.79e-02 & \cellcolor[RGB]{143,204,163} 4.78e-03 \\
 & GPT-4o & \cellcolor[RGB]{216,232,178} 9.88e-04 & \cellcolor[RGB]{158,210,166} 2.31e-04 & \cellcolor[RGB]{249,244,185} 1.67e-02 & \cellcolor[RGB]{199,225,175} 2.31e-02 & \cellcolor[RGB]{144,205,163} 4.88e-03 \\
 & o3 & \cellcolor[RGB]{216,232,178} 9.88e-04 & \cellcolor[RGB]{168,214,168} 2.74e-04 & \cellcolor[RGB]{244,157,90} 2.03e-01 & \cellcolor[RGB]{145,205,164} 1.49e-02 & \cellcolor[RGB]{143,204,163} 4.81e-03 \\
\midrule
\multirow{6}{*}{\textbf{CodePDE}}
 & Gemma 3 & \cellcolor[RGB]{239,127,91} 5.61e-03 & \cellcolor[RGB]{236,103,92} 5.17e-02 & \cellcolor[RGB]{243,155,90} 2.13e-01 & \cellcolor[RGB]{246,176,92} 9.29e-02 & \cellcolor[RGB]{248,202,128} 7.69e-02 \\
 & LLaMA 3.3 & \cellcolor[RGB]{236,105,92} 7.37e-03 & \cellcolor[RGB]{237,107,92} 4.59e-02 & \cellcolor[RGB]{243,154,90} 2.18e-01 & \cellcolor[RGB]{237,110,92} 2.36e-01 & \cellcolor[RGB]{236,102,92} 1.03e+00 \\
 & Qwen 3 & \cellcolor[RGB]{241,138,91} 4.89e-03 & \cellcolor[RGB]{250,235,173} 1.35e-03 & \cellcolor[RGB]{236,102,92} 9.55e-01 & \cellcolor[RGB]{236,103,92} 2.59e-01 & \cellcolor[RGB]{238,119,91} 6.57e-01 \\
 & DeepSeek-R1 & \cellcolor[RGB]{218,233,179} 1.01e-03 & \cellcolor[RGB]{175,216,170} 3.04e-04 & \cellcolor[RGB]{243,155,90} 2.13e-01 & \cellcolor[RGB]{221,234,180} 2.80e-02 & \cellcolor[RGB]{143,204,163} 4.80e-03 \\
 & GPT-4o & \cellcolor[RGB]{250,230,165} 1.55e-03 & \cellcolor[RGB]{186,220,172} 3.65e-04 & \cellcolor[RGB]{250,240,179} 1.99e-02 & \cellcolor[RGB]{240,128,91} 1.81e-01 & \cellcolor[RGB]{238,119,91} 6.57e-01 \\
 & o3 & \cellcolor[RGB]{214,231,178} 9.74e-04 & \cellcolor[RGB]{168,214,168} 2.74e-04 & \cellcolor[RGB]{250,240,179} 1.99e-02 & \cellcolor[RGB]{246,176,92} 9.29e-02 & \cellcolor[RGB]{144,205,163} 4.88e-03 \\
\midrule
\multirow{6}{*}{\textbf{CodePDE-R}}
 & Gemma 3 & \cellcolor[RGB]{243,150,90} 4.20e-03 & \cellcolor[RGB]{247,191,113} 4.63e-03 & \cellcolor[RGB]{245,164,89} 1.69e-01 & \cellcolor[RGB]{248,203,129} 6.44e-02 & \cellcolor[RGB]{249,223,156} 4.47e-02 \\
 & LLaMA 3.3 & \cellcolor[RGB]{220,233,179} 1.02e-03 & \cellcolor[RGB]{247,191,113} 4.59e-03 & \cellcolor[RGB]{245,169,89} 1.43e-01 & \cellcolor[RGB]{237,110,92} 2.36e-01 & \cellcolor[RGB]{245,167,89} 1.92e-01 \\
 & Qwen 3 & \cellcolor[RGB]{214,231,178} 9.74e-04 & \cellcolor[RGB]{185,220,172} 3.60e-04 & \cellcolor[RGB]{247,185,105} 9.13e-02 & \cellcolor[RGB]{159,210,167} 1.67e-02 & \cellcolor[RGB]{144,205,163} 4.90e-03 \\
 & DeepSeek-R1 & \cellcolor[RGB]{218,233,179} 1.01e-03 & \cellcolor[RGB]{176,217,170} 3.15e-04 & \cellcolor[RGB]{249,244,185} 1.67e-02 & \cellcolor[RGB]{159,210,167} 1.67e-02 & \cellcolor[RGB]{143,204,163} 4.80e-03 \\
 & GPT-4o & \cellcolor[RGB]{214,231,178} 9.74e-04 & \cellcolor[RGB]{165,213,168} 2.57e-04 & \cellcolor[RGB]{249,244,185} 1.67e-02 & \cellcolor[RGB]{201,226,175} 2.36e-02 & \cellcolor[RGB]{143,204,163} 4.80e-03 \\
 & o3 & \cellcolor[RGB]{218,233,179} 1.01e-03 & \cellcolor[RGB]{185,220,172} 3.60e-04 & \cellcolor[RGB]{245,169,89} 1.43e-01 & \cellcolor[RGB]{131,200,161} 1.31e-02 & \cellcolor[RGB]{144,205,163} 4.90e-03 \\
\midrule
\multirow{6}{*}{\textbf{PDE-SHARP}}
 & Gemma 3 & \cellcolor[RGB]{218,233,179} 1.01e-03 & \cellcolor[RGB]{211,230,177} 5.60e-04 & \cellcolor[RGB]{147,206,164} 3.01e-03 & \cellcolor[RGB]{235,239,183} 3.14e-02 & \cellcolor[RGB]{225,235,180} 1.72e-02 \\
 & LLaMA 3.3 & \cellcolor[RGB]{217,232,179} 9.98e-04 & \cellcolor[RGB]{200,226,175} 4.61e-04 & \cellcolor[RGB]{158,210,166} 3.61e-03 & \cellcolor[RGB]{249,220,152} 5.06e-02 & \cellcolor[RGB]{225,235,180} 1.72e-02 \\
 & Qwen 3 & \cellcolor[RGB]{183,219,172} 7.76e-04 & \cellcolor[RGB]{173,216,170} 2.97e-04 & \cellcolor[RGB]{131,200,161} 2.32e-03 & \cellcolor[RGB]{221,234,180} 2.80e-02 & \cellcolor[RGB]{143,204,163} 4.80e-03 \\
 & DeepSeek-R1 & \cellcolor[RGB]{131,200,161} 5.24e-04 & \cellcolor[RGB]{131,200,161} 1.48e-04 & \cellcolor[RGB]{131,200,161} 2.29e-03 & \cellcolor[RGB]{136,201,162} 1.37e-02 & \cellcolor[RGB]{143,204,163} 4.74e-03 \\
 & GPT-4o & \cellcolor[RGB]{151,207,165} 6.11e-04 & \cellcolor[RGB]{158,210,166} 2.31e-04 & \cellcolor[RGB]{131,200,161} 2.29e-03 & \cellcolor[RGB]{148,206,164} 1.51e-02 & \cellcolor[RGB]{131,200,161} 3.97e-03 \\
 & o3 & \cellcolor[RGB]{214,231,178} 9.74e-04 & \cellcolor[RGB]{182,219,171} 3.42e-04 & \cellcolor[RGB]{186,220,172} 5.78e-03 & \cellcolor[RGB]{175,216,170} 1.89e-02 & \cellcolor[RGB]{173,216,170} 7.78e-03 \\
\midrule
\bottomrule
\end{tabular}
  }
\end{table}

\paragraph{PDE-SHARP is more robust to code generator LLM selection.}
Table \ref{tab:nrmse-results} shows that the solution quality for baseline methods depends strongly on the LLM.
In contrast, PDE-SHARP performs more consistently across all tested LLMs; results for more LLMs are appear Appendix \ref{Results with Different LLMs}. This uniform performance indicates PDE-SHARP's tournament hybridization stage effectively mitigates the limitations of individual code generators, producing higher-quality solvers that are largely independent of the underlying LLM. 

\begin{figure}[h]
    \centering
    \begin{subfigure}[b]{0.48\textwidth}
        \centering
        \includegraphics[width=\textwidth]{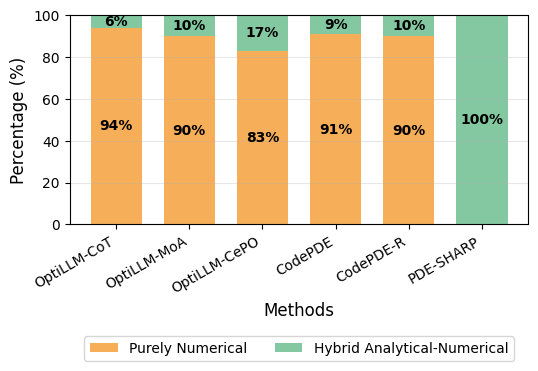}
        \caption{Reaction-Diffusion}
        \label{fig:react-diff}
    \end{subfigure}
    \hfill
    \begin{subfigure}[b]{0.48\textwidth}
        \centering        \includegraphics[width=\textwidth]{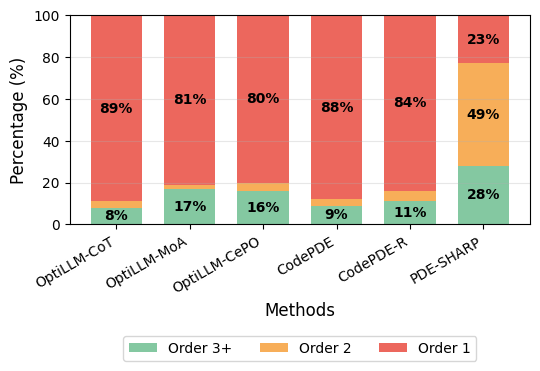}
        \caption{Advection}
        \label{fig:advection}
    \end{subfigure}

    \caption{(a) 
    Other frameworks tend to choose the less accurate purely-numerical approach for the reaction-diffusion PDE, while PDE-SHARP always goes with the superior hybrid approach.  (b) PDE-SHARP transitions from first-order discretized analytical to second-order finite-volume approaches through performance-informed tournaments.}
    \label{advect_react_diff_percentages}
\end{figure}

\paragraph{PDE-SHARP significantly improves solver accuracy for specific PDEs.} 
PDE-SHARP improves accuracy by over $4 \times$ overall (geometric mean), with particularly impressive performance on the reaction-diffusion and advection tasks. For reaction-diffusion, PDE-SHARP's Analysis stage immediately identifies that the reaction component admits an analytical solution, directing all 32 initial solver candidates toward hybrid analytical-numerical approaches that achieve superior numerical stability. Baseline methods rarely discover this hybrid strategy, as shown in Figure \ref{fig:react-diff}.

For advection, PDEBench reference solutions are generated using finite volume methods~\citep{takamoto2024pdebenchextensivebenchmarkscientific}, reflecting standard shock-safe computational practice. 
PDE-SHARP and all other baselines initially attempt analytical solutions, and the baselines keep their analytical approach even through refinement (e.g. in CodePDE-R). PDE-SHARP's performance-informed tournaments, on the other hand, encourage PDE-SHARP to adapt to the data, as demonstrated in Figure \ref{fig:advection}. When persistent $10^{-3}$ errors reported as feedback indicate a mismatch between analytical and reference solutions, the judge LLMs converge on second-order finite-volume schemes that better match the dataset characteristics. This adaptation occurs through feedback alone, without manual intervention, demonstrating how collaborative tournaments can optimize for evaluation criteria while maintaining computational efficiency. This adaptive behavior varies with different feedback types as users can choose an optimization target to reflect available data (Figure \ref{advection nrmse and no feedback}). 
A study on advection solvers appears in Appendix~\ref{Case Study on Advection}.

\begin{figure}[h!]
    \centering
\includegraphics[width=\linewidth]{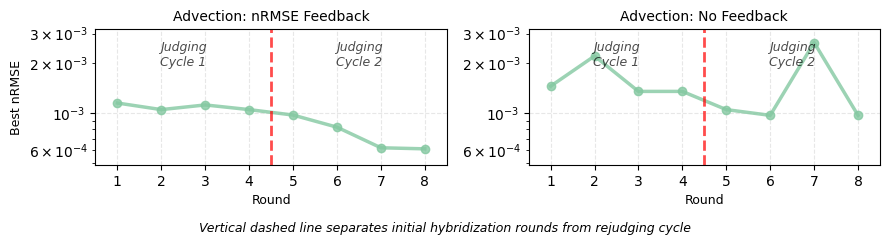}
    \caption{Without proper feedback, the judges stick to analytical approaches. Figure \ref{PDE-SHARP num rounds ablation advection} gives details. }
    \label{advection nrmse and no feedback}
\end{figure}

\begin{figure}[h!]
    \centering
\includegraphics[width=0.6\linewidth]{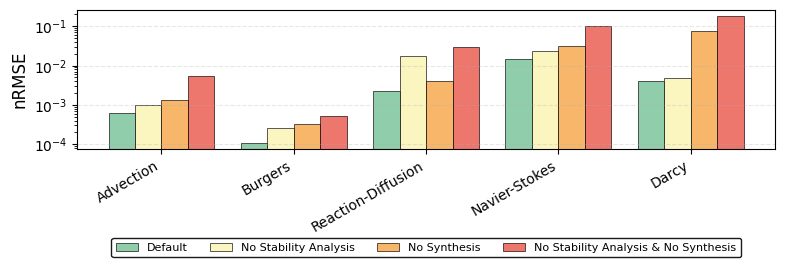}
   \caption{Ablation study of PDE-SHARP components  across five PDE tasks. Four variants: (1) Default: full PDE-SHARP with both stability analysis and synthesis, (2) No Stability Analysis: PDE-SHARP with the stability analysis step removed from the Analysis stage, (3) No Synthesis: PDE-SHARP with best-of-32 sampling instead of the Synthesis stage, and (4) No Stability Analysis \& No Synthesis. Results show both components contribute to accuracy improvements, with each component being more critical for different PDE types, e.g. stability analysis is more critical for reaction-diffusion, while synthesis contributes more to the Darcy flow task.}
\label{fig:stability and tournaments effects}
\end{figure}

Figures \ref{advect_react_diff_percentages}, \ref{advection nrmse and no feedback}, \ref{fig:stability and tournaments effects}  highlight how PDE-SHARP's \textbf{Analysis} and \textbf{Synthesis} stages leverage mathematical insight and performance feedback, both playing significant roles in PDE-SHARP's performance. Detailed ablation studies in Appendix \ref{PDE-SHARP Ablation Studies} quantify each component's contribution in more detail.

\subsubsection{Case Study Summary: Reaction–Diffusion Solver Evolution}

A detailed account of this synthesis process appears in Appendix~\ref{Example of a Synthesis Stage Tournament Report: Reaction-Diffusion PDE Solver Refinement}. Here, we summarize a report of the evolution of a top-performing solver for the 1D reaction–diffusion PDE
\[
\partial_t u - \nu \partial_{xx} u - \rho u(1 - u) = 0 \quad \text{with periodic boundaries on } x \in (0,1),
\]
under a parameter regime of $\nu = 0.5$ and $\rho = 1.0$ during PDE-SHARP's hybrid tournaments. The process involved four rounds of judge-guided iteration and hybridization, and ultimately reduced the L2 error by a factor of 77$\times$—from 0.166 to 0.002—without changing the time step size or switching to implicit methods.

\vspace{1mm}
\textbf{Round 1 (Baseline):} The initial solver used Lie splitting (reaction followed by diffusion), where the reaction was integrated analytically via a logistic formula. While the implementation was correct and stable, its time step was overly conservative due to a misapplied stability constraint that included both diffusion and reaction terms. Since the reaction step is exact (not discretized), its inclusion unnecessarily limited $\Delta t$, resulting in over two million substeps per output trajectory and moderate L2 error (0.166).

\vspace{1mm}
\textbf{Round 2 (Hybridization Attempt):} To address inefficiencies, judges merged features from other candidate codes: using only the diffusion-based $\Delta t$ bound, switching to Strang splitting for second-order accuracy, and retaining the analytical reaction. However, these changes failed to reduce cost or improve accuracy. The $\Delta t$ bound remained governed by diffusion due to spatial resolution ($N=1024$), and the increased number of operator applications in Strang splitting amplified phase errors. The L2 error worsened slightly (0.185), despite using the same number of internal steps.

\vspace{1mm}
\textbf{Round 3 (Implicit Diffusion):} Attempting a more substantial architectural change, the judges replaced explicit diffusion with an unconditionally stable Crank–Nicolson method. This permitted much larger time steps (up to 39$\times$ larger), but performance degraded sharply (L2 error 0.301). Investigation revealed that the tridiagonal solver was incompatible with periodic boundary conditions and that the simplified Lie splitting strategy (one reaction step followed by one implicit diffusion step per output time) created significant truncation error. The attempted optimization thus failed due to both boundary mismatch and operator imbalance.

\vspace{1mm}
\textbf{Round 4 (Targeted Local Fix):} Judges reverted to the more promising Strang splitting configuration from Round 2, but focused on a subtle issue in the reaction step. Although mathematically correct, the original logistic integration formula suffered from catastrophic floating-point cancellation: when $u \approx 0$ or $u \approx 1$, the formula introduced numerical instability or overflow. The key breakthrough was a numerically stable reformulation of the same expression:
\[
\texttt{return } 1 / \left(1 + e^{-\rho \Delta t} \cdot \frac{1 - u}{u + \epsilon}\right), \quad \epsilon = 10^{-10}.
\]
This stabilized the computation without altering the underlying algorithm. The result was a dramatic improvement: L2 error dropped to 0.002 while keeping the same time step size and structure as the original baseline. This final solver was robust, accurate, and production-quality—free of NaNs, Infs, or edge-case failures.

\vspace{1mm}
\textbf{Key Lessons:}
\begin{itemize}
    \item \emph{Numerical stability often outweighs algorithmic complexity.} The most impactful change was not in the solver structure, but in improving a single line of floating-point math.
    \item \emph{Blind use of advanced methods can harm performance.} The implicit method in Round 3 degraded accuracy due to mismatched assumptions and poor boundary treatment.
    \item \emph{Incremental, analysis-driven refinement is highly effective.} Careful error tracing and domain-specific insight enabled major improvements with minimal code changes.
\end{itemize}

\subsection{Code Quality \& Insights}
\begin{figure}[h]
    \centering
    \begin{subfigure}[b]{0.28\textwidth}
        \centering \includegraphics[width=\textwidth]{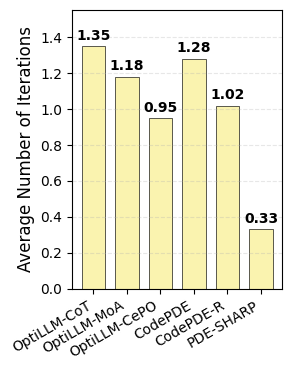}
        \caption{}
        \label{fig:debugging}
    \end{subfigure}
    \hfill
    \begin{subfigure}[b]{0.7\textwidth}
        \centering        \includegraphics[width=\textwidth]{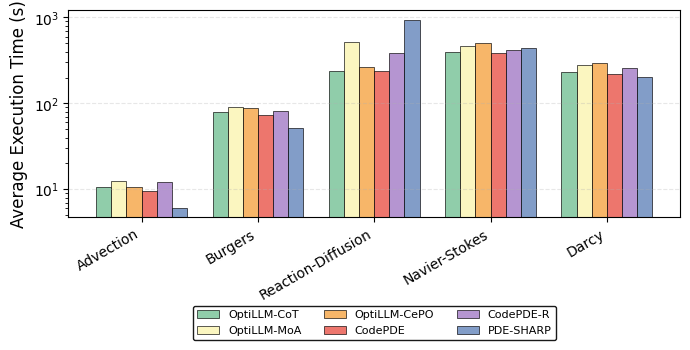}
        \caption{}
        \label{fig:all runtimes}
    \end{subfigure}

    \caption{(a) Average number of debugging iterations required per solver execution across different methods. (b) Average execution times across PDE tasks. PDE-SHARP achieves lower execution times than the average baseline in 4/5 cases. For reaction-diffusion, higher execution time reflects the rigorous numerical methods selected by stability analysis as expected, which produce significantly higher accuracy solvers (Table \ref{tab:nrmse-results}). }
    \label{debugging_execution_time}
\end{figure}

Figure \ref{debugging_execution_time} demonstrates PDE-SHARP reduces the number of debugging iterations required and produces solvers with competitive execution times.
PDE-SHARP averages 0.33 debugging iterations per solver execution (approximately 1 in 3 generated solvers requires debugging in a hybridization round), significantly outperforming baseline methods that require 0.9--1.4 debugging iterations per generated solver. This reduction shows that PDE-SHARP's Analysis stage produces more robust initial implementations, and that the synthesis stage efficiently eliminates implementation errors.

 Figure \ref{fig:advection} demonstrates the distribution of empirical convergence orders (definition in Appendix \ref{Additional Experimental Setups}) --- showing solver improvements with grid refinement --- across methods for the advection PDE. PDE-SHARP generates solvers with superior convergence properties, leading to higher accuracy in this case (Table \ref{tab:nrmse-results}). In addition, Figure \ref{fig:solver-library-usage-grid} indicates that on average, PDE-SHARP's solvers use less PyTorch (down to $\approx$25--33 \% of library calls) and more SciPy + NumPy + JAX (up to $\approx$60--75 \%), whereas the baselines keep PyTorch at roughly 50--67\% and SciPy below 7\% on average. Using JAX for computational kernels is highly encouraged in PDE-SHARP prompts in particular as evident in the library usage proportions across all methods and PDE tasks. Additional empirical convergence rate results all PDEs as well as library usage proportions for each baseline appear in Appendix \ref{Analysis of the Generated Solver Code Quality}.

\begin{figure}[h!]
    \centering
\includegraphics[width=\linewidth]{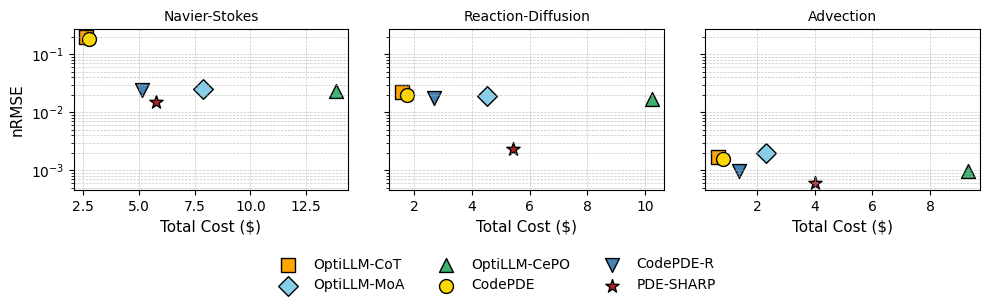}
    \caption{Trade-off between solution accuracy (nRMSE) and total cost for LLM-driven PDE solver generation methods across three PDE tasks of varying computational complexity. From Navier-Stokes (hours per solver evaluation) to Reaction-Diffusion (moderate) to Advection (lightweight, seconds per evaluation), PDE-SHARP demonstrates consistent cost-effectiveness.}
    \label{nrmse vs cost}
\end{figure}

\paragraph{Cost.}  We analyze the efficiency and cost of each method by calculating the average cost for GPU and LLM API calls for the experiments in this section. Table \ref{tab:nrmse-results} shows among the tested LLMs, GPT-4o as the code generation LLM yields higher accuracy results on average.  Table \ref{framework_costs} shows the total average API cost of the results for GPT-4o in Table \ref{tab:nrmse-results}. Details of the calculations appear in Appendix \ref{Additional Information on Framework Cost}. GPU usage depends on the number of solver executions, code complexity, and implementation efficiency. 
The number of solver executions for PDE-SHARP depends on the number of hybridization rounds required, averaging 13.2 evaluations across all test cases (9-12 evaluations for most PDEs, with advection requiring 24 to better match data as discussed in Section \ref{Results and Analysis}). Figure \ref{nrmse vs cost} shows nRMSE vs. total average cost (API call + GPU usage) for three PDE tasks.

\begin{table}[htbp]
  \centering
  \caption{Cost comparison of input, output, and total API usage per method using GPT-4o as the code generating LLM.}
  \label{framework_costs}
  \begin{tabular}{l r r r r}
    \toprule
    \textbf{Framework} & \textbf{\$ Inputs} & \textbf{\$ Output} & \textbf{\$ Total} \\
    \midrule
    OptiLLM-CoT                    & 0.10    & 0.48   & 0.58  \\
    OptiLLM-MoA                    & 0.53    & 2.12   & 2.65  \\
    OptiLLM-CePO                   & 0.96   & 8.27   &  9.23 \\
    CodePDE & 0.07 & 0.68  & 0.75 \\
    CodePDE-R & 0.41 & 0.88  & 1.29  \\
    PDE-SHARP                         & 1.12    & 2.89  & 4.01  \\
    \bottomrule
  \end{tabular}
\end{table}

\subsection{Discussion \& Limitations}

\textbf{Discussion:} PDE-SHARP uses numerical feedback to improve the generated solver. This extra information can be easy to compute --- such as the (data-free) PDE residual --- or may require collecting data, such as distance to the solution at a sampled set of times and locations. PDE-SHARP can also use problem-specific constraints like the CFL condition~\citep{leveque2007finite} as feedback, or can run without feedback if no information is available. Results for PDE-SHARP using residual feedback and no numerical feedback appear in Appendix \ref{PDE-SHARP Ablation Studies}. LLM agents can also suggest feedback types. As seen in Appendix \ref{Hybridization Feedback Type} (examples of LLM-suggested feedback types for each tested PDE), an additional LLM agent could be used to determine optimal problem-specific metrics before Synthesis begins. This is particularly beneficial for complex PDEs requiring specialized feedback, and represents important future work. Additional promising directions include scaling to higher-dimensional problems with complex geometries where traditional numerical methods face greater challenges. Finally, hybrid approaches combining PDE-SHARP's interpretable numerical solvers with neural PDE methods could leverage the strengths of both paradigms for problems requiring both accuracy and computational efficiency.

\textbf{Limitations:}
Our evaluation establishes PDE-SHARP's effectiveness on moderate-complexity PDEs from established benchmarks, with high-fidelity computational simulations representing a natural extension constrained by current LLM training data coverage. LLM-driven PDE solver generation using test-time computing approaches rely on LLM mathematical reasoning capabilities, which means performance may degrade for cutting-edge PDE formulations that are not well-represented in training data or require highly specialized domain knowledge beyond current model capabilities.


 
\section{Conclusion}

PDE-SHARP demonstrates that intelligent LLM-driven solver generation can dramatically improve efficiency over brute-force sampling approaches. Our three-stage framework reduces computational evaluations by 60-75\% while achieving superior accuracy on average across five representative PDEs. The mathematical chain-of-thought analysis in the Analysis stage produces more robust initial implementations, requiring on average 67\% fewer debugging iterations compared to baseline methods. The hybrid tournaments in the Synthesis stage efficiently refines solvers through performance-informed feedback, with flexible type, demonstrating consistent robust improvements across diverse LLM models.

\bibliography{references}
\bibliographystyle{iclr2026_conference}
\clearpage
\appendix
\section*{Appendix}
\renewcommand{\thesection}{\Alph{section}}

\section{Additional Information \& Experimental Setups} \label{Additional Experimental Setups}

\subsection{Mathematical Metrics}
\paragraph{nRMSE.} For $S$ test cases, each with true solution $u^{(s)}(x,t)$ and solver prediction $\hat{u}^{(s)}(x,t)$:

$$\text{nRMSE} = \frac{1}{S} \sum_{s=1}^{S} \frac{\| u^{(s)}(x,t) - \hat{u}^{(s)}(x,t) \|_2}{\| u^{(s)}(x,t) \|_2}$$

where $\|\cdot\|_2$ denotes the L2 norm. This metric normalizes the root mean squared error by the magnitude of the true solution, enabling fair comparison across problems with different solution scales.
\paragraph{Convergence Rate.} To evaluate numerical correctness, we assess solver convergence behavior across multiple grid resolutions. A robust solver should exhibit predictable error reduction following $E(h) \approx Ch^p$, where $E(h)$ is the solution error on grid spacing $h$, $C$ is a problem-dependent constant, and $p$ is the convergence order.

We estimate the empirical convergence order using two grid resolutions:
$$p \approx \frac{\log\left(\frac{E(h_1)}{E(h_2)}\right)}{\log\left(\frac{h_1}{h_2}\right)}$$

For each generated solver, we evaluate performance on progressively refined grids (typically $h$, $h/2$, $h/4$) and compute the average convergence order. Expected theoretical orders vary by numerical method: first-order schemes ($p \approx 1$), second-order finite difference/volume methods ($p \approx 2$), and spectral methods (exponential convergence). Most LLM-generated solvers achieve first-order convergence, with occasional higher-order behavior depending on the chosen discretization scheme and implementation quality.




\subsection{Neural Networks \& Foundation Models}

\paragraph{Limitations of Cross-Paradigm Comparisons.} Direct comparison between LLM-generated solvers using traditional numerical methods and neural PDE solvers involves inherent methodological challenges. Neural network baselines are drawn from prior literature with different experimental conditions while our LLM approach benefits from extensive inference-time optimization (debugging, refinement, best-of-n sampling) not applied to these baselines. Additionally, the computational budgets differ fundamentally: neural methods require training time and data preparation, while numerical methods require implementation and parameter tuning effort. These paradigmatic differences make it difficult to establish truly equivalent experimental conditions. Our results should be interpreted as demonstrating the promise of LLM-based solver generation rather than definitive superiority over alternative approaches. Future work should focus on controlled comparisons with standardized evaluation protocols across all methods.

We thus include the following reported numbers verbatim from the original papers of FNO \citep{li2020fno}, PirateNets \citep{wang2024piratenets}, PINNsFormer \citep{zhao2023pinnsformer}, and UPS \citep{shen2024upsefficientlybuildingfoundation} as examples of neural and foundation  models only for the sake of completeness and to give readers an at-a-glance sense of scale (parameters, memory, time/epoch) and accuracy on overlapping PDE families, however, as each method utilizes distinct settings, we do not provide a direct ranking between them. The following is intended only to document the resource scale and the published accuracy on broadly overlapping PDE families.

\vspace{0.75ex}
\noindent\textbf{FNO} Reports results for 1D Burgers and 2D Navier--Stokes (space--time operator learning). Hardware noted by the authors: single NVIDIA V100 16\,GB.

\begin{table}[h]
\centering
\caption{FNO on \emph{1D Burgers} (relative $\ell_2$ error at different spatial resolutions $s$).}
\label{tab:fno-burgers}
\begin{tabular}{lcccccc}
\toprule
Method & $s{=}256$ & $512$ & $1024$ & $2048$ & $4096$ & $8192$\\
\midrule
FNO & 0.0149 & 0.0158 & 0.0160 & 0.0146 & 0.0142 & 0.0139\\
\bottomrule
\end{tabular}
\vspace{0.25ex}

\footnotesize\emph{Notes.} Table reproduced from the paper; parameters, GPU memory, and time/epoch were not reported for the Burgers experiment. See Table~\ref{tab:fno-ns} for Navier--Stokes resource numbers as reported by the authors.
\end{table}

\begin{table}[h]
  \centering
  \caption{FNO on \emph{2D Navier--Stokes} (relative $\ell_2$ error over different viscosities $\nu$ and dataset sizes $N$; per‑epoch time reported by the authors).}
  \label{tab:fno-ns}
  \resizebox{\textwidth}{!}{%
    \begin{tabular}{lcccccc}
      \toprule
      Method & Params & Time/epoch & $\nu{=}10^{-3}$, $T{=}50$, $N{=}1000$ & $\nu{=}10^{-4}$, $T{=}30$, $N{=}1000$ & $\nu{=}10^{-4}$, $T{=}30$, $N{=}10000$ & $\nu{=}10^{-5}$, $T{=}20$, $N{=}1000$ \\
      \midrule
      FNO-3D  & 6{,}558{,}537 & 38.99\,s  & 0.0086 & 0.1918 & 0.0820 & 0.1893 \\
      FNO-2D  &   414{,}517   & 127.80\,s & 0.0128 & 0.1559 & 0.0973 & 0.1556 \\
      U\,-Net & 24{,}950{,}491 & 48.67\,s & 0.0245 & 0.2051 & 0.1190 & 0.1982 \\
      TF\,-Net& 7{,}451{,}724 & 47.21\,s  & 0.0225 & 0.2253 & 0.1168 & 0.2268 \\
      ResNet  &   266{,}641   & 78.47\,s  & 0.0701 & 0.2871 & 0.2311 & 0.2753 \\
      \bottomrule
    \end{tabular}%
  }
\footnotesize\emph{Notes.} Reported at $64{\times}64$ spatial resolution; FNO-3D convolves in space--time while FNO-2D uses 2D convolutions with an RNN in time.  
\end{table}

\vspace{1ex}
\noindent\textbf{PirateNets} has PINN backbone with physics-informed residual adaptive blocks. The paper emphasizes accuracy comparisons and ablations; it does not tabulate parameter counts, GPU memory, or wall-clock per epoch. Below we list the state-of-the-art test errors the authors report.

\begin{table}[h]
\centering
\caption{PirateNets: reported relative $\ell_2$ test errors across PDEs (paper’s Table~1).}
\label{tab:pirate-sota}
\begin{tabular}{lcccc}
\toprule
Benchmark & Error (PirateNet) & Params & GPU Mem & Time/epoch\\
\midrule
Allen--Cahn (1D) & $2.24{\times}10^{-5}$ & --- & --- & ---\\
Korteweg--De Vries (1D) & $4.27{\times}10^{-4}$ & --- & --- & ---\\
Grey--Scott (2D) & $3.61{\times}10^{-3}$ & --- & --- & ---\\
Ginzburg--Landau (2D) & $1.49{\times}10^{-2}$ & --- & --- & ---\\
Lid-driven cavity (2D) & $4.21{\times}10^{-2}$ & --- & --- & ---\\
\bottomrule
\end{tabular}

\footnotesize\emph{Notes.} Architecture details (e.g., depth/width) and training pipelines are provided, but resource metrics are not tabulated.
\end{table}

\vspace{1ex}
\noindent\textbf{PINNsFormer} is a transformer-style PINN variant. The authors report parameter counts and training overhead (V100), and test errors on overlapping 1D PDEs.

\begin{table}[h!]
\centering
\caption{PINNsFormer: model size and training overhead (Appendix Table~4–5 in the paper).}
\label{tab:pinnsformer-overhead}
\begin{tabular}{lccc}
\toprule
Model & Params & GPU Mem (MiB) & Time/epoch (s)\\
\midrule
PINNsFormer (pseudo-seq.\ length $k{=}5$) & 454{,}000 & 2{,}827 & 2.34\\
\bottomrule
\end{tabular}

\footnotesize\emph{Notes.} Reported on a single NVIDIA Tesla V100; overheads shown for $k{=}5$.
\end{table}

\begin{table}[h!]
\centering
\caption{PINNsFormer: reported test errors on 1D PDEs used widely in PINN literature.}
\label{tab:pinnsformer-accuracy}
\begin{tabular}{lcccc}
\toprule
PDE (dimension) & Metric (paper) & Error & Params & Time/epoch / GPU Mem\\
\midrule
Convection (1D) & rRMSE ($\approx$ rel.\ $\ell_2$) & 0.027 & 454k & 2.34\,s / 2{,}827\,MiB\\
Reaction (1D)   & rRMSE ($\approx$ rel.\ $\ell_2$) & 0.030 & 454k & 2.34\,s / 2{,}827\,MiB\\
\bottomrule
\end{tabular}

\footnotesize\emph{Notes.} Errors are taken directly from the paper’s main results tables; rRMSE is the paper’s standard relative $\ell_2$ metric. The reaction/convection formulations and sampling follow the setups specified in \cite{zhao2023pinnsformer}.
\end{table}

\vspace{1ex}
\noindent\textbf{UPS} learns to map symbolic PDE specifications and initial/boundary conditions to numerical solutions. The architecture combines Fourier Neural Operators and transformers with autoregressive decoding over space-time grids.

The model was trained on $\sim$20k PDE trajectories using a single NVIDIA A6000 GPU. Training was run for 60{,}000 steps and completed in under 100 GPU-hours. UPS achieves strong sample efficiency, outperforming baselines with 4$\times$ less data and 26$\times$ less compute.

\begin{table}[h]
\centering
\caption{UPS: test errors on PDEBench benchmarks (relative $\ell_2$ or nRMSE as reported).}
\label{tab:ups-pdebench}
\begin{tabular}{lccccc}
\toprule
PDE & Metric & Error (UPS) & Training Steps & GPU & Total GPU Hours \\
\midrule
Advection (1D) & nRMSE & $2.20{\times}10^{-3}$ & 60{,}000 & A6000 & <100 \\
Burgers (1D) & nRMSE & $3.73{\times}10^{-2}$ & 60{,}000 & A6000 & <100 \\
Reaction–Diffusion (2D) & nRMSE & $5.57{\times}10^{-2}$ & 60{,}000 & A6000 & <100 \\
Navier–Stokes (2D) & nRMSE & $4.50{\times}10^{-3}$ & 60{,}000 & A6000 & <100 \\
\bottomrule
\end{tabular}

\footnotesize\emph{Notes.} Errors and training configuration are from the paper’s PDEBench experiments. Training used $\sim$20k PDE samples across equations; GPU time and steps refer to total training, not per-PDE.
\end{table}

\subsection{LLM-Driven Architectures}
\subsubsection{LLM Models Used in  Section 
\ref{Experiments} for Code Generation}
\label{LLM Models From Experiments}
\begin{table}[htbp]
  \centering
  \caption{LLM models used in Section \ref{Experiments} for solver generation; more LLMs -- including the coding and math‑aware variants of these -- are tested in Appendix~\ref{Results with Different LLMs}}
  \label{main_lls}
  \begin{tabular}{l l l}
    \toprule
    \textbf{LLM} & \textbf{Type} & \textbf{Access} \\
    \midrule
    Gemma 3    & Non-reasoning & Open Source \\
    LLama 3.3     & Non‑reasoning   & Open Source \\
    Qwen3         & Non‑reasoning   & Open Source \\
    DeepSeek‑R1   & Reasoning       & Open Source \\
    GPT‑4o        & Non‑reasoning   & API Service   \\
    o3            & Reasoning       & API Service   \\
    \bottomrule
  \end{tabular}
\end{table}

\subsubsection{Agentic Workflows}

Frameworks like FunSearch ~\citep{FunSearch2023} and AIDE ~\citep{jiang2025aideaidrivenexplorationspace} wrap an LLM in an iterative search/refinement loop. They treat the LLM as an agent that can branch, try multiple approaches, and refine code via feedback.

\textbf{FunSearch (DeepMind, 2023)}  pairs a pre-trained code-generating LLM with an automated evaluator in a loop. The LLM proposes candidate programs/solutions, an evaluator (a test or objective function) checks them, and then the process generates new candidates (mutations, combinations) based on feedback. 
FunSearch features algorithm discovery based on a program database. The program
database consists of a few “islands” of programs. The experimental setup is the same as ~\citep{li2025codepdeinferenceframeworkllmdriven}. The number of islands is set to 4
and the island reset period to 3600s. The FunSearch process runs for 32 iterations. In each iteration, the
language model decoding temperature is set to 0.7.

\textbf{AIDE (Weco AI, 2025) } formulates code generation as a tree search problem. For a given high-level task (like “build an ML pipeline that achieves X accuracy on Y dataset”), AIDE would have the LLM propose a solution. Then it measures how good that solution is (it runs the code and sees accuracy). If not satisfied, AIDE can either refine the current solution (edit some parts of the code via another LLM call) or try a different approach (branch out in the search tree). Over multiple iterations, it explores the space of programs.  The experimental setup is the same as ~\citep{li2025codepdeinferenceframeworkllmdriven}.  AIDE runs for 96 steps and the max debug depth, debug probability, and number of drafts are set to 5, 0.9, and 24,
respectively. The language model decoding temperature is set to 0.5 for code generation following the original paper ~\citep{jiang2025aideaidrivenexplorationspace}.

\begin{table}[!ht]
\centering
\caption{\small nRMSE values for Agentic Workflows on different PDEs. Results from \cite{li2025codepdeinferenceframeworkllmdriven}}
\label{tab:nrmse-agentic}
\vspace{0.3cm}
\resizebox{0.9\linewidth}{!}{%
\begin{tabular}{c ccccc}
\toprule
 & \textbf{Advection} & \textbf{Burgers} & \textbf{Reaction-Diffusion} & \textbf{Navier-Stokes} & \textbf{Darcy}\\
\midrule
AIDE & 1.03\text{e}-3 & 1.05\text{e}-4 & 5.07\text{e}-2 & 5.77\text{e}-2 & 4.78\text{e}-3 \\
FunSearch & 1.05\text{e}-3 & 1.13\text{e}-4 & 3.72\text{e}-2 & 5.86\text{e}-2 & 4.78\text{e}-3 \\
\bottomrule
\end{tabular}
}
\vspace{-1mm}
\end{table}

\subsubsection{Other Related Work} \label{Results from Related Work}

Recent work ~\cite{soroco2025pdecontrollerllmsautoformalizationreasoning} introduces PDE-Controller, a framework that fine-tunes LLMs specifically for PDE control problems. Their approach trains specialized models for autoformalization (converting natural language to formal specifications), program synthesis, and multi-step reasoning through reinforcement learning from human feedback (RLHF). While demonstrating strong performance on their target domains, this approach differs from PDE-SHARP in several key aspects.

\begin{table}[h]
\centering
\caption{PDE-Controller: Training Requirements and Performance}
\label{tab:pde_controller_stats}
\begin{tabular}{lr}
\toprule
\textbf{Metric} & \textbf{Value} \\
\midrule
\textbf{Training Data} & \\
Heat equation samples & 867,408 \\
Wave equation samples & 845,088 \\
Total training samples & 1,712,496 \\
\midrule
\textbf{Evaluation Data} & \\
Synthetic test samples & 426,432 \\
Manual test problems & 34 \\
\midrule
\textbf{Performance (Synthetic)} & \\
Autoformalization accuracy (IoU) & 99.2\% \\
Code executability & 97.99\% \\
\midrule
\textbf{Performance (Manual)} & \\
Autoformalization accuracy (IoU) & 68.0\% \\
Code executability & 91.2\% \\
\midrule
\textbf{Scope} & \\
PDE types covered & 2 (heat, wave) \\
Spatial dimensions & 1D  \\
\bottomrule
\end{tabular}
\end{table}

While effective for specific classes of PDEs, the fine-tuning approach presents several limitations compared to LLM-driven approaches using test-time computing: \textbf{(1) Computational overhead:} Requires extensive fine-tuning of multiple specialized models (translator, controller, coder) with over 1.7M training samples; \textbf{(2) Domain specificity:} Limited to only heat and wave equations in 1D, requiring retraining for new PDE types; \textbf{(3) Data requirements:} Needs large-scale synthetic data generation and manual curation by domain experts; \textbf{(4) Scalability constraints:} Each new PDE family would require collecting new training data and retraining models; \textbf{(5) Generalization gap:} Performance drops significantly on manual problems (99.2\% to 68.0\% accuracy), indicating limited robustness to real-world variations.

PDE-SHARP offers more flexibility across PDE types without domain-specific training, though potentially at the cost of specialized performance on specific equation families. The fundamental trade-off lies between the specialized efficiency of fine-tuned approaches versus the broader applicability and reduced computational overhead of general prompting strategies.

\subsubsection{OptiLLM}
We use the OptiLLM framework from \texttt{github.com/codelion/optillm} as a baseline to test PDE-SHARP. OptiLLM is an optimizing inference proxy that implements 20+ state-of-the-art techniques to improve LLM accuracy and performance on reasoning tasks without requiring any model training or fine-tuning. We test three of OptiLLM's implemented techniques in our study.

\paragraph{CoT (Chain-of-Thought) with Reflection.} Implements chain-of-thought reasoning with structured $<$thinking$>$, $<$reflection$>$ and $<$output$>$ sections to enhance reasoning quality through explicit self-evaluation. The approach generates intermediate reasoning steps in the thinking phase, critically reviews the reasoning in the reflection phase, and produces the final output, enabling improved accuracy on complex reasoning tasks without requiring model fine-tuning.

\paragraph{MoA (Mixture-of-Agents).} Combines responses from multiple model critiques in a collaborative framework where 3 different agent perspectives are aggregated to produce higher-quality solutions. 

\paragraph{CePO (Cerebras Planning and Optimization).} Combines Best-of-$n$ sampling (without code execution), Chain-of-Thought reasoning, Self-Reflection, and Self-Improvement in a four-stage process: plan generation with confidence scoring, initial solution development, plan refinement through inconsistency analysis, and final solution production. The method applies Best-of-$n$ to multiple solution candidates with optional plan diversity, using parameters like \texttt{planning\_n} proposals and \texttt{planning\_m} maximum attempts to generate robust solutions for complex reasoning tasks. The following are the default parameters used in this study.

\begin{table}[h!]
  \centering
  \caption{Default configuration values for CePO planning and verification stages}
  \label{cepo_config_defaults}
  \resizebox{\textwidth}{!}{%
  \begin{tabular}{l l l}
    \toprule
    \textbf{Parameter} & \textbf{Description} & \textbf{Default Value} \\
    \midrule
    \texttt{--cepo\_bestofn\_n} & Number of responses to be generated in best of n stage & 3 \\
    \texttt{--cepo\_bestofn\_temperature} & Temperature for verifier in best of n stage & 0.1 \\
    \texttt{--cepo\_bestofn\_max\_tokens} & Max tokens for verifier in best of n stage & 4096 \\
    \texttt{--cepo\_bestofn\_rating\_type} & Rating type ("absolute" or "pairwise") & \texttt{"absolute"} \\
    \texttt{--cepo\_planning\_n} & Number of plans generated in planning stage & 3 \\
    \texttt{--cepo\_planning\_m} & Attempts to generate \texttt{n} plans in planning stage & 6 \\
    \texttt{--cepo\_planning\_temperature\_step1} & Temperature in step 1 of planning stage & 0.55 \\
    \texttt{--cepo\_planning\_temperature\_step2} & Temperature in step 2 of planning stage & 0.25 \\
    \texttt{--cepo\_planning\_temperature\_step3} & Temperature in step 3 of planning stage & 0.1 \\
    \texttt{--cepo\_planning\_temperature\_step4} & Temperature in step 4 of planning stage & 0 \\
    \texttt{--cepo\_planning\_max\_tokens\_step1} & Max tokens in step 1 of planning stage & 4096 \\
    \texttt{--cepo\_planning\_max\_tokens\_step2} & Max tokens in step 2 of planning stage & 4096 \\
    \texttt{--cepo\_planning\_max\_tokens\_step3} & Max tokens in step 3 of planning stage & 4096 \\
    \texttt{--cepo\_planning\_max\_tokens\_step4} & Max tokens in step 4 of planning stage & 4096 \\
    \texttt{--cepo\_print\_output} & Whether to print the output of each stage & \texttt{False} \\
    \texttt{--cepo\_config\_file} & Path to CePO configuration file & \texttt{None} \\
    \texttt{--cepo\_use\_plan\_diversity} & Use additional plan diversity step & \texttt{False} \\
    \texttt{--cepo\_rating\_model} & Rating model (if different from completion) & \texttt{None} \\
    \bottomrule
  \end{tabular}%
  }
\end{table}

\subsubsection{CodePDE}

\paragraph{CodePDE.} CodePDE~\citep{li2025codepdeinferenceframeworkllmdriven} is an inference framework for LLM-driven PDE solver generation that frames PDE solving as a code generation task. The framework operates through a five-step process: (1) \textit{Task Specification} converts PDE problems into natural language descriptions including governing equations, domain specifications, boundary conditions, and initial conditions; (2) \textit{Code Generation} uses chain-of-thought prompting to instruct models to generate complete solver implementations with predefined function signatures; (3) \textit{Debugging} performs iterative self-debugging for up to 4 rounds when solvers encounter execution errors, feeding error traces back to the LLM for autonomous correction; and (4) \textit{Evaluation} assesses solver performance using normalized root mean squared error (nRMSE), convergence tests, and execution time; For our comparison, we use CodePDE with the same setup as ~\citep{li2025codepdeinferenceframeworkllmdriven} with steps 1-4 (reasoning + debugging), generating 32 solver samples with best-of-32 selection, using up to 4 debugging iterations per solver.

\paragraph{CodePDE-R.} CodePDE-R extends the base CodePDE framework by incorporating the solver refinement step (step 5). This variant selects the 5 best-performing programs from the reasoning + debugging stage as "seed" programs for refinement. The refinement process provides the nRMSE obtained during evaluation along with the solver implementation back to the LLM, instructing it to analyze execution results, identify numerical instabilities and bottlenecks, and generate improved implementations accordingly. For each seed program, the framework generates 4 refined versions across different refinement configurations (using 3, 4, or 5 seed implementations), resulting in 12 refined programs total. The final result reports the best nRMSE among these 12 refined samples. This iterative feedback-driven optimization enables models to systematically improve solver accuracy and efficiency beyond the initial generation and debugging phases.

\subsection{Additional Information on Framework Cost} \label{Additional Information on Framework Cost}

Table \ref{framework_costs} shows the average API call cost for each framework using GPT-4o as the code generator LLM. GPT-4o input cost is $\$2.50$ per $1$M tokens, and the output cost is  $\$10.00$ per $1$M tokens. Table \ref{num tokens} shows the average input-output counts for each framework from Section \ref{Experiments}. An NVIDIA T4 GPU costs $\$0.35$ per hour, which is used to calculate the total average costs in Figure \ref{nrmse vs cost}.

\begin{table}[h!]
  \centering
  \caption{Approximation of the total input-output counts for running each framework once}
  \label{framework_io_counts}
  \begin{tabular}{l r r}
    \toprule
    \textbf{Framework} & \textbf{\# Inputs} & \textbf{\# Output} \\
    \midrule
    OptiLLM (CoT)    & 48,000        & 105,600   \\
    OptiLLM (MoA)    & 200,000       & 422,400   \\
    OptiLLM (CePO)   & 600,000       & 105,600   \\
    CodePDE          & 102,400       & 294,400   \\
    PDE-SHARP         & 600,000 & 450,800   \\
    \bottomrule
  \end{tabular}
  \label{num tokens}
\end{table}

\newpage
\section{Additional Experimental Results} \label{Additional Experimental Results}

\subsection{Results with Different LLMs} \label{Results with Different LLMs}

The following additional LLM models are tested for code generation in addition to the results of Table \ref{tab:nrmse-results}.

\begin{table}[h!]
  \centering
  \caption{Additional LLMs}
  \label{llm_capabilities}
  \begin{tabular}{l l l}
    \toprule
    \textbf{LLM} & \textbf{Type} & \textbf{Access} \\
    \midrule
    Qwen3‑Coder ~\citep{qwen3technicalreport} & Coding-specific & Open Source \\
    Code Llama ~\citep{rozière2024codellamaopenfoundation} & Coding-specific & Open Source \\
    GPT-5 & Non-reasoning & API Service \\
    DeepSeekMath ~\citep{shao2024deepseekmathpushinglimitsmathematical} & Mathematical reasoning & Open Source \\
    DeepSeek-Coder ~\citep{guo2024deepseekcoderlargelanguagemodel} & Coding-specific & Open Source \\
    MathCoder2-DeepSeekMath ~\citep{lu2024mathcoder2bettermathreasoning} & Math aware Coding-specific & Open Source \\
    \bottomrule
  \end{tabular}
\end{table}

\begin{table}[h!]
  \centering
  \caption{nRMSE comparison of the baseline frameworks using different LLMs.}
  \label{tab:other-llms-nrmse-results}
  \resizebox{\textwidth}{!}{%
\begin{tabular}{llccccc}
\toprule
 & & Advection & Burgers & Reaction-Diffusion & Navier-Stokes & Darcy \\
\midrule
\multirow{5}{*}{\textbf{OptiLLM-CoT}}
 & Qwen3-Coder & 4.67e-03 & 1.52e-03 & 9.38e-01 & 2.63e-01 & 6.34e-01 \\
 & GPT-5 & 5.36e-03 & 1.88e-03 & 1.04e+00 & 2.83e-01 & 7.18e-01 \\
 & DeepSeekMath & 4.89e-03 & 3.12e-04 & 2.38e-01 & 8.51e-02 & 5.22e-03 \\
 & DeepSeek-Coder & 4.89e-03 & 3.04e-04 & 2.41e-01 & 8.72e-02 & 5.11e-03 \\
 & MathCoder2-DeepSeekMath & 4.89e-03 & 3.27e-04 & 2.43e-01 & 8.66e-02 & 5.29e-03 \\
\midrule
\multirow{5}{*}{\textbf{OptiLLM-MoA}}
 & Qwen3-Coder & 1.01e-03 & 3.45e-04 & 9.68e-02 & 1.79e-02 & 5.12e-03 \\
 & GPT-5 & 4.18e-03 & 4.11e-04 & 1.14e-01 & 2.02e-02 & 1.89e-02 \\
 & DeepSeekMath & 1.32e-03 & 2.66e-04 & 3.57e-02 & 1.72e-02 & 5.23e-03 \\
 & DeepSeek-Coder & 1.32e-03 & 3.04e-04 & 1.55e-01 & 1.78e-02 & 5.18e-03 \\
 & MathCoder2-DeepSeekMath & 1.01e-03 & 2.66e-04 & 4.07e-02 & 1.74e-02 & 5.22e-03 \\
\midrule
\multirow{5}{*}{\textbf{OptiLLM-CePO}}
 & Qwen3-Coder & 1.01e-03 & 3.23e-04 & 8.91e-02 & 1.97e-02 & 1.83e-02 \\
 & GPT-5 & 3.17e-03 & 3.89e-04 & 1.03e-01 & 2.24e-02 & 4.72e-02 \\
 & DeepSeekMath & 9.98e-04 & 2.55e-04 & 2.45e-02 & 1.85e-02 & 4.92e-03 \\
 & DeepSeek-Coder & 1.01e-03 & 2.66e-04 & 1.47e-01 & 1.91e-02 & 4.92e-03 \\
 & MathCoder2-DeepSeekMath & 9.98e-04 & 3.04e-04 & 3.56e-02 & 1.93e-02 & 4.33e-03 \\
\midrule
\multirow{5}{*}{\textbf{CodePDE}}
 & Qwen3-Coder & 4.89e-03 & 1.35e-03 & 9.55e-01 & 2.59e-01 & 6.57e-01 \\
 & GPT-5 & 5.75e-03 & 1.63e-03 & 1.08e-01 & 2.82e-01 & 7.91e-01 \\
 & DeepSeekMath & 5.10e-03 & 2.87e-04 & 2.45e-02 & 7.91e-02 & 4.97e-03 \\
 & DeepSeek-Coder & 4.69e-03 & 2.87e-04 & 2.78e-01 & 7.82e-02 & 5.02e-03 \\
 & MathCoder2-DeepSeekMath & 5.10e-03 & 3.15e-04 & 2.32e-02 & 7.84e-02 & 4.97e-03 \\
\midrule
\multirow{5}{*}{\textbf{CodePDE-R}}
 & Qwen3-Coder & 9.74e-04 & 3.60e-04 & 9.13e-02 & 9.67e-02 & 4.90e-02 \\
 & GPT-5 & 1.14e-03 & 4.41e-04 & 1.07e-01 & 7.93e-02 & 5.81e-02 \\
 & DeepSeekMath & 9.89e-04 & 2.62e-04 & 1.47e-02 & 3.63e-02 & 5.01e-03 \\
 & DeepSeek-Coder & 9.89e-04 & 3.15e-04 & 1.47e-02 & 2.67e-02 & 6.01e-03 \\
 & MathCoder2-DeepSeekMath & 9.74e-04 & 2.62e-04 & 1.47e-02 & 1.65e-02 & 4.97e-03 \\
\midrule
\multirow{5}{*}{\textbf{PDE-SHARP}}
 & Qwen3-Coder & 9.74e-04 & 2.97e-04 & 5.39e-03 & 2.80e-02 & 7.80e-03 \\
 & GPT-5 & 1.01e-03 & 3.45e-04 & 7.78e-03 & 3.19e-02 & 9.93e-03 \\
 & DeepSeekMath & 7.46e-04 & 1.55e-04 & 2.39e-03 & 1.47e-02 & 4.78e-03 \\
 & DeepSeek-Coder & 7.46e-04 & 2.53e-04 & 3.67e-03 & 2.76e-02 & 4.78e-03 \\
 & MathCoder2-DeepSeekMath & 5.54e-04 & 1.38e-04 & 2.99e-03 & 1.47e-02 & 3.93e-03 \\
\midrule
\bottomrule
\end{tabular}
}
\end{table}

\subsection{PDE-SHARP Ablation Studies} \label{PDE-SHARP Ablation Studies}
In this section, we present ablation study results on PDE-SHARP. Note that we take the default PDE-SHARP framework to be one used in Section \ref{Experiments}. The ablation studies of this section each target a different aspect of PDE-SHARP's design.

\subsubsection{Analysis Prompting Strategy} \label{Analysis Prompting Strategy}

We compare the following prompting strategies for the Analysis stage. 
\begin{itemize}
    \item Multi-Step prompting (PDE-SHARP default)
    \item Single Prompt (all the PDE-SHARP steps merged into one)
    \item LLM-generated multi-step prompting
    \item LLM-generated single prompt
\end{itemize}

For the LLM-generated alternatives, the LLM, GPT-4o in this ablation, is first asked to generate either a series of prompts or a single prompt to run as the analysis stage for a give PDE before proceeding to the code generation stage. The Synthesis stage is done exactly as in Section \ref{Experiments}. Table \ref{tab:ablation-analysis-results} summarizes these results. 

\begin{table}[h!]
  \centering
  \caption{nRMSE comparison of the baseline frameworks using different Analysis prompting strategies.}
  \label{tab:ablation-analysis-results}
  \resizebox{\textwidth}{!}{%

\begin{tabular}{llccccc}
\toprule
 & & Advection & Burgers & Reaction-Diffusion & Navier-Stokes & Darcy \\
\midrule
\multirow{5}{*}{\textbf{Multi-Step Prompting (Default)}}
 & Gemma 3 & 1.01e-03 & 5.60e-04 & 3.01e-03 & 3.14e-02 & 1.72e-02 \\
 & LLaMA 3.3 & 9.98e-04 & 4.61e-04 & 3.61e-03 & 5.06e-02 & 1.72e-02 \\
 & Qwen 3 & 7.76e-04 & 2.97e-04 & 2.32e-03 & 2.80e-02 & 4.80e-03 \\
 & DeepSeek-R1 & 5.24e-04 & 1.48e-04 & 2.29e-03 & 1.37e-02 & 4.74e-03 \\
 & GPT-4o & 6.11e-04 & 2.31e-04 & 2.29e-03 & 1.51e-02 & 3.97e-03 \\
 & o3 & 9.74e-04 & 3.42e-04 & 5.78e-03 & 1.89e-02 & 7.78e-03 \\
\midrule
\multirow{5}{*}{\textbf{Single Prompt (Default merged into one)}}
 & Gemma 3 & 1.03e-03 & 4.89e-04 & 1.18e-02 & 4.31e-02 & 8.11e-03 \\
 & LLaMA 3.3 & 1.05e-03 & 4.79e-04 & 1.75e-02 & 7.32e-02 & 1.79e-02 \\
 & Qwen 3 & 8.01e-04 & 3.11e-04 & 2.41e-03 & 4.94e-02 & 4.91e-03 \\
 & DeepSeek-R1 & 6.53e-04 & 1.56e-04 & 2.37e-03 & 1.41e-02 & 4.83e-03 \\
 & GPT-4o & 7.39e-04 & 3.48e-04 & 3.33e-03 & 2.62e-02 & 4.13e-03 \\
 & o3 & 8.70e-04 & 4.54e-04 & 3.89e-03 & 2.96e-02 & 4.87e-03 \\
\midrule
\multirow{5}{*}{\textbf{LLM-Generated Multi-Step Prompting}}
 & Gemma 3 & 1.02e-03 & 4.82e-04 & 9.21e-02 & 7.27e-02 & 7.93e-03 \\
 & LLaMA 3.3 & 1.04e-03 & 4.72e-04 & 8.69e-02 & 7.24e-02 & 1.77e-02 \\
 & Qwen 3 & 1.89e-03 & 6.05e-04 & 3.39e-02 & 3.89e-02 & 4.85e-03 \\
 & DeepSeek-R1 & 8.37e-04 & 5.30e-04 & 1.33e-02 & 3.40e-02 & 4.85e-03 \\
 & GPT-4o & 7.27e-04 & 4.15e-04 & 1.31e-02 & 2.59e-02 & 4.05e-03 \\
 & o3 & 6.96e-04 & 7.48e-04 & 1.84e-02 & 3.93e-02 & 4.85e-03 \\
\midrule
\multirow{5}{*}{\textbf{LLM-Generated Single Prompt}}
 & Gemma 3 & 1.04e-03 & 4.95e-04 & 1.29e-01 & 5.42e-02 & 8.19e-03 \\
 & LLaMA 3.3 & 1.06e-03 & 6.87e-04 & 1.81e-01 & 6.43e-02 & 1.81e-02 \\
 & Qwen 3 & 1.13e-03 & 6.19e-04 & 8.47e-02 & 3.98e-02 & 3.95e-03 \\
 & DeepSeek-R1 & 9.59e-04 & 4.95e-04 & 1.39e-02 & 4.43e-02 & 4.85e-03 \\
 & GPT-4o & 2.47e-03 & 7.22e-04 & 2.36e-02 & 3.65e-02 & 4.85e-03 \\
 & o3 & 9.19e-04 & 7.48e-04 & 3.91e-02 & 3.01e-02 & 5.92e-03 \\
\midrule
\bottomrule
\end{tabular}
}
\end{table}

\begin{figure}[H]
    \centering
\includegraphics[width=0.6\linewidth]{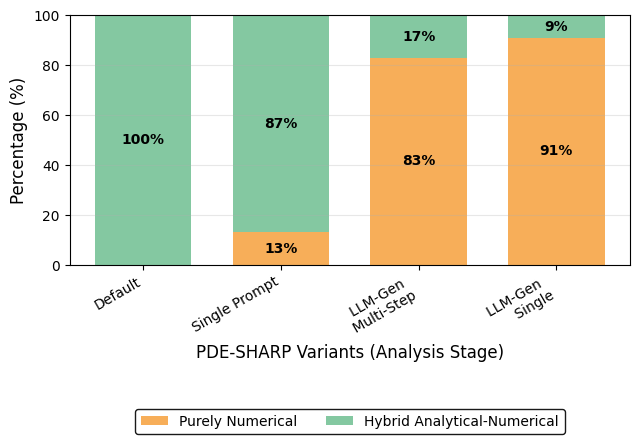}
    \caption{Solver strategy selection for reaction-diffusion PDE across PDE-SHARP variants. LLM-generated prompts do not usually lead to optimal solver strategy selection in this case.}
    \label{stage 1 variants react diff}
\end{figure}

Our experiments demonstrate that the Multi-Step Prompting strategy consistently yields the best performance across all LLMs and PDEs. When all the PDE-SHARP Analysis prompts are merged together into a single prompt, LLMs tend to not follow the instructions thoroughly as they become too long to follow. Moreover, when the LLM is tasked with generating the prompts for the analysis stage, it is observed that many details, such as checking for hybrid approaches or doing a rigorous numerical stability analysis is overlooked. Analyzing the strategies used in the generated solvers (Table \ref{tab:ablation-analysis-results}) for the reaction-diffusion task is a great demonstration of this shortcoming as reaction diffusion is more sensitive to method choice and stability analysis (Figure \ref{stage 1 variants react diff}). Naturally, the most pronounced impact is observed on the Reaction-Diffusion PDE, where the default multi-step approach achieves the lowest average nRMSE of 2.88e-03 across all LLMs. In contrast, the average nRMSE rises to 6.88e-03 with Single Prompting, 4.30e-02 with LLM-Generated Multi-Step Prompting, and peaks at 7.86e-02 with LLM-Generated Single Prompting. This corresponds to a 27$\times$ increase in error from the best case to the worst, highlighting the critical role of well-structured multi-step analysis in improving solution accuracy for complex PDEs.

\subsubsection{The Effects of Stability Analysis} \label{The Effects of Stability Analysis} 

\begin{figure}[h!]
    \centering
\includegraphics[width=0.95\linewidth]{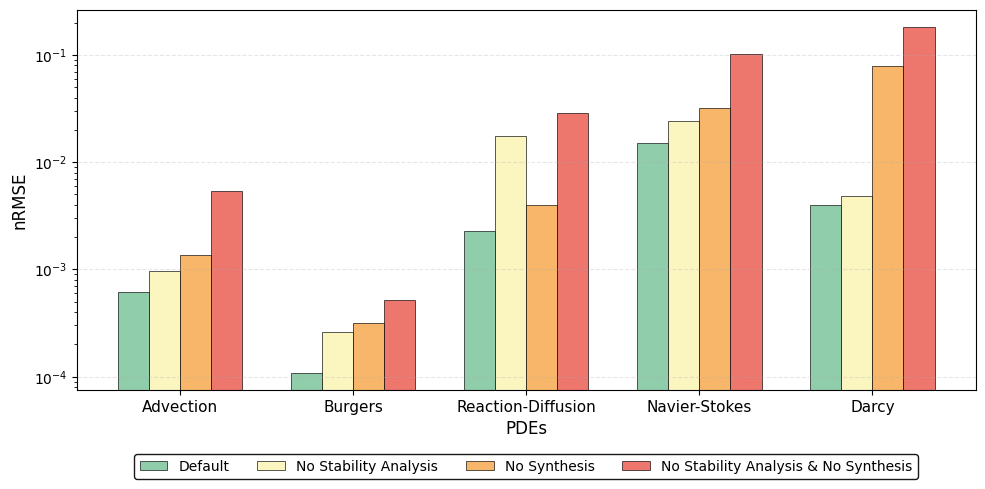}
    \caption{Ablation study of PDE-SHARP components across five PDE tasks. Results show that mathematical stability analysis is critical for solver accuracy, while tournaments provide additional improvements. Removing both components leads to significant performance degradation, particularly for complex PDEs like Darcy flow.}
    \label{PDE-SHARP stability ablation}
\end{figure}

To evaluate the individual contributions of PDE-SHARP's key components --- the stability analysis in the Analysis stage and the tournaments in the Synthesis stage --- we conduct an ablation study examining four variants: (1) the default framework with both mathematical stability analysis and tournaments, (2) tournaments without stability analysis, (3) stability analysis without tournaments (best-of-32 sampling with stability analysis), and (4) neither component (best-of-32 sampling without stability analysis). Figure \ref{PDE-SHARP stability ablation} demonstrates that mathematical stability analysis provides substantial accuracy improvements across all tested PDEs. Removing stability analysis while maintaining tournaments increases average nRMSE by 2-8$\times$ depending on the PDE complexity. The tournaments component shows mixed but generally positive effects, with the largest improvements observed for reaction-diffusion and Darcy flow problems. Most critically, removing both components results in significant performance degradation, with nRMSE increases of 5-45$\times$ for complex PDEs like Darcy flow. These results confirm that PDE-SHARP's mathematical analysis stage is essential for generating numerically stable solvers, while the tournament-based refinement provides additional accuracy gains particularly for challenging nonlinear problems.

\begin{figure}[H]
    \centering
\includegraphics[width=0.6
\linewidth]{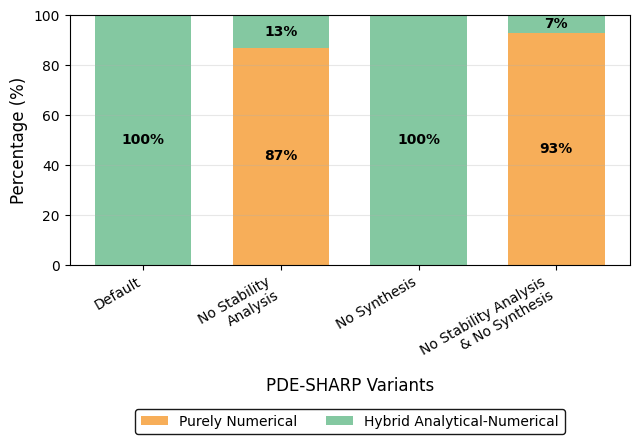}
    \caption{Solver strategy selection for reaction-diffusion PDE across PDE-SHARP variants. Mathematical stability analysis (present in Default and No Tournaments variants) consistently guides the framework toward superior hybrid analytical-numerical approaches, while its absence leads to predominantly numerical methods.}
    \label{PDE-SHARP stability ablation react-diff}
\end{figure}

The stability analysis component of PDE-SHARP plays a crucial role in guiding solver strategy selection. Figure \ref{PDE-SHARP stability ablation react-diff} illustrates the percentage of hybrid analytical-numerical versus purely numerical approaches chosen by each PDE-SHARP variant for the reaction-diffusion equation. The default framework and the variant without tournaments both achieve 100\% hybrid approach selection, demonstrating that mathematical stability analysis consistently identifies the superiority of hybrid methods for this PDE. In contrast, removing stability analysis results in predominantly numerical approaches (87-93\%), as the framework lacks the mathematical insight to recognize that the reaction component admits an analytical solution. This strategic difference directly explains the accuracy improvements observed in the previous ablation study, as hybrid approaches achieve superior numerical stability and precision for reaction-diffusion problems.

\subsubsection{Reasoning vs. Non-reasoning LLMs For Code Generation In \textbf{Genesis}} 

Experiments indicate that in PDE-SHARP, there is negligible difference between the final results using reasoning, non-reasoning, coding-specific, and mathematical LLM models (Tables \ref{main_lls} \& \ref{llm_capabilities}) as the code generator in the Genesis stage. See Tables \ref{tab:nrmse-results} and \ref{tab:other-llms-nrmse-results} for nRMSE results.

\subsubsection{Test-time Scaling for PDE-SHARP} \label{Test-time Scaling for PDE-SHARP}

\begin{figure}[h]
    \centering
    \includegraphics[width=\linewidth]{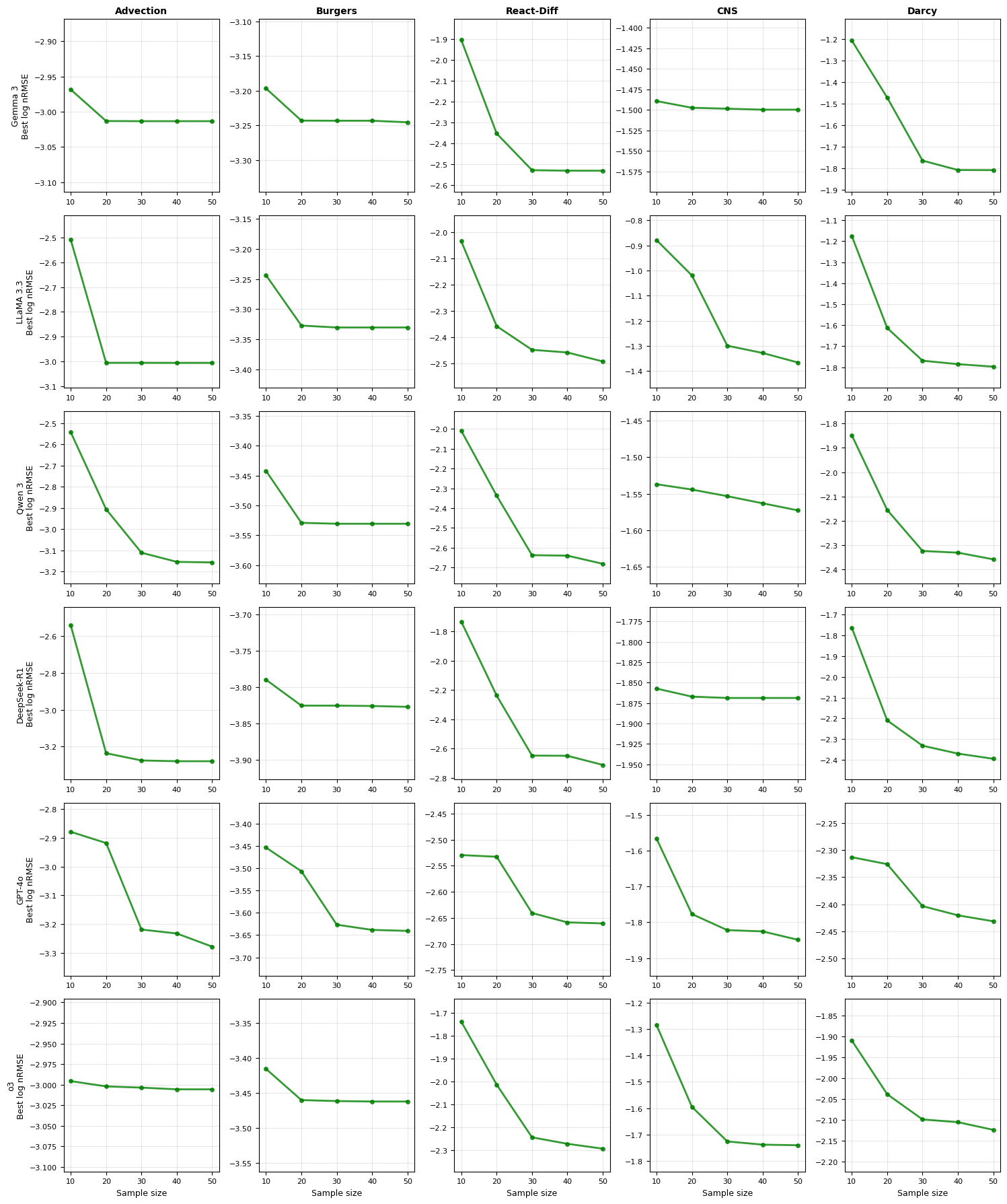}
    \caption{Varying the number of solver samples generated for each LLM and each PDE family in PDE-SHARP.}
    \label{fig:test time PDE-SHARP}
\end{figure}

 Based on our test-time scaling study (Figure \ref{fig:test time PDE-SHARP}) for PDE-SHARP and to be consistent with findings from ~\citep{li2025codepdeinferenceframeworkllmdriven} on the same PDE tasks, we use $n=32$ initial solver candidates in our experiments. This choice balances computational efficiency with sufficient diversity for effective solver selection in the subsequent Synthesis stage.

\newpage

\subsubsection{Structure of the Tournaments} \label{Structure of the Tournaments}

In this ablation study, we keep the default PDE-SHARP strategy from Section \ref{Experiments} for the Analysis and Genesis stages and replace the Synthesis stage with various strategies to study its effectiveness. In PDE-SHARP's default Synthesis stage in Section \ref{Experiments}, three LLM instances, which we call "judges", are tasked with the selection and hybridization tournaments. To achieve the best performance (Table \ref{tab:nrmse-results}) --- i.e. fewer tournament rounds to get the highest performing PDE solver codes --- these three judges are taken to be a mixture of reasoning and non-reasoning LLMs (o3, DeepSeek-R1, and GPT-4o) in Section \ref{Experiments}. This set of LLM judges are chosen to balance efficient code generation and code stability details with the detailed reasoning and attention to numerical implementation details that the reasoning models bring in. In this section, we consider other possibilities for the three judges to justify our choice of LLM judges.  Tables \ref{tab:nrmse-results} and \ref{tab:other-llms-nrmse-results} demonstrate that using different LLM models to generate 32 samples of solver codes leads to overall negligible difference in the final results in PDE-SHARP as the tournaments lead to solvers robust to LLM choice. Thus, we stick to the default GPT-4o for code generation in this ablation study and use the same 32 samples generated by GPT-4o for all of the stage 3 strategies studied. Note that in these tournaments, feedback type is set to be nRMSE similar to Section \ref{Experiments}. Results for different feedback types are presented later in this section. Since numerous LLM configurations exist, we select a minimal representative subset from each category. Current models have sufficient input capacity for tournament solver lists; future work could incorporate summarizer agents to compress information for smaller models.

We test six tournament structure categories:

{1. Mixed Judges (Default):} Combines reasoning and non-reasoning models to balance code generation efficiency with detailed numerical reasoning:
\begin{itemize}
\item o3 + GPT-4o + DeepSeek-R1 (Section \ref{Experiments} default)
\item o3 + GPT-4o + GPT-4o
\item DeepSeek-R1 + GPT-4o + GPT-4o
\end{itemize}

{2. All Reasoning Judges:} Uses only reasoning-capable models:
\begin{itemize}
\item o3 + o3 + o3
\item DeepSeek-R1 + DeepSeek-R1 + DeepSeek-R1
\item o3 + o3 + DeepSeek-R1
\end{itemize}

{3. All Non-Reasoning Judges:} Uses only standard language models:
\begin{itemize}
\item GPT-4o + GPT-4o + GPT-4o
\end{itemize}

{4. Best-of-32 Baseline:} Executes all 32 solvers from Analysis and Genesis stages without tournaments.

{5. Fixed Criteria Judging:} Applies categories 1-3 with predetermined evaluation criteria:
\begin{itemize}
\item Numerical stability and convergence properties
\item Computational efficiency and scalability
\item Mathematical correctness and precision
\item Implementation robustness and error handling
\item Solution accuracy on benchmark problems
\end{itemize}

{6. Self-Generated Criteria:} Applies categories 1-3 where judges first generate their own evaluation criteria before selection.

All strategies use identical 32 solver samples from GPT-4o code generation to ensure fair comparison.

\newpage

\begin{table}[h!]
  \centering
  \caption{nRMSE values for each PDE-SHARP using different LLM combinations for the Synthesis stage.}
  \label{tab:syntheis_ablation}
  \resizebox{\textwidth}{!}{%
\begin{tabular}{llccccc}
\toprule
 & & Advection & Burgers & Reaction-Diffusion & Navier-Stokes & Darcy \\
\midrule
\multirow{3}{*}{\textbf{Mixed Judges (Default)}}
 & o3 + GPT-4o + DeepSeek-R1 & 6.11e-04 & 2.31e-04 & 2.29e-03 & 1.51e-02 & 3.97e-03 \\
 & o3 + GPT-4o + GPT-4o & 7.34e-04 & 4.45e-04 & 5.41e-03 & 3.58e-02 & 4.12e-03 \\
 & DeepSeek-R1 + GPT-4o + GPT-4o & 6.98e-04 & 2.31e-04 & 4.33e-03 & 1.51e-02 & 4.91e-03 \\
\midrule
\multirow{3}{*}{\textbf{All Reasoning}}
 & o3 + o3 + o3 & 9.74e-04 & 5.19e-04 & 4.21e-03 & 3.45e-02 & 3.84e-03 \\
 & DeepSeek-R1 + DeepSeek-R1 + DeepSeek-R1 & 8.92e-04 & 3.23e-04 & 3.25e-03 & 2.47e-02 & 3.84e-03 \\
 & o3 + o3 + DeepSeek-R1 & 7.79e-04 & 2.35e-04 & 4.33e-03 & 1.51e-02 & 3.97e-03 \\
\midrule
\multirow{1}{*}{\textbf{All Non-Reasoning}}
 & GPT-4o + GPT-4o + GPT-4o & 9.74e-04 & 2.57e-04 & 1.01e-02 & 2.62e-02 & 4.90e-03 \\
\midrule
\multirow{1}{*}{\textbf{Best-of-32 Baseline}}
 & No Tournaments & 1.35e-03 & 3.19e-04 & 3.99e-03 & 3.18e-02 & 7.82e-02 \\
\midrule
\multirow{3}{*}{\textbf{Fixed Criteria - Mixed Judges}}
 & o3 + GPT-4o + DeepSeek-R1 & 9.86e-04 & 5.25e-04 & 7.24e-03 & 1.48e-02 & 3.89e-03 \\
 & o3 + GPT-4o + GPT-4o & 9.18e-04 & 2.38e-04 & 2.36e-02 & 1.54e-02 & 4.05e-03 \\
 & DeepSeek-R1 + GPT-4o + GPT-4o & 1.01e-03 & 2.21e-04 & 8.27e-03 & 1.46e-02 & 3.85e-03 \\
\midrule
\multirow{3}{*}{\textbf{Fixed Criteria - All Reasoning}}
 & o3 + o3 + o3 & 1.73e-03 & 6.11e-04 & 1.15e-02 & 1.41e-02 & 7.76e-03 \\
 & DeepSeek-R1 + DeepSeek-R1 + DeepSeek-R1 & 9.74e-04 & 3.17e-04 & 3.19e-03 & 1.44e-02 & 3.82e-03 \\
 & o3 + o3 + DeepSeek-R1 & 1.68e-03 & 2.08e-04 & 1.12e-02 & 2.89e-02 & 3.73e-03 \\
\midrule
\multirow{1}{*}{\textbf{Fixed Criteria - All Non-Reasoning}}
 & GPT-4o + GPT-4o + GPT-4o & 1.01e-03 & 3.43e-04 & 2.42e-03 & 9.29e-02 & 5.01e-03 \\
\midrule
\multirow{3}{*}{\textbf{Self-Generated Criteria - Mixed Judges}}
 & o3 + GPT-4o + DeepSeek-R1 & 8.12e-04 & 4.67e-04 & 9.15e-03 & 1.62e-02 & 4.21e-03 \\
 & o3 + GPT-4o + GPT-4o & 8.53e-04 & 3.02e-04 & 1.89e-02 & 1.38e-02 & 4.57e-03 \\
 & DeepSeek-R1 + GPT-4o + GPT-4o & 1.15e-03 & 2.94e-04 & 6.83e-03 & 1.71e-02 & 3.42e-03 \\
\midrule
\multirow{3}{*}{\textbf{Self-Generated Criteria - All Reasoning}}
 & o3 + o3 + o3 & 1.58e-03 & 7.24e-04 & 1.38e-02 & 1.27e-02 & 8.35e-03 \\
 & DeepSeek-R1 + DeepSeek-R1 + DeepSeek-R1 & 9.13e-04 & 2.85e-04 & 4.06e-03 & 1.59e-02 & 4.18e-03 \\
 & o3 + o3 + DeepSeek-R1 & 1.52e-03 & 2.76e-04 & 9.84e-03 & 2.53e-02 & 4.29e-03 \\
\midrule
\multirow{1}{*}{\textbf{Self-Generated Criteria - All Non-Reasoning}}
& GPT-4o + GPT-4o + GPT-4o & 9.27e-04 & 3.89e-04 & 3.17e-03 & 8.46e-02 & 5.68e-03 \\
\midrule
\bottomrule
\end{tabular}
  }
\end{table}

\begin{table}[h!]
  \centering
  \caption{Number of rounds to achieve the results of Table \ref{tab:syntheis_ablation} for each PDE-SHARP using different LLM combinations for the Synthesis stage. The number of rounds is reported before performance saturation/degradation,  indicating the minimum number of hybridization rounds. The ``+" sign indicates a rejudging cycle as explained in Table \ref{num rounds}. Note that no hybrid tournaments accur in the best-of-32 strategy.}
  \label{tab:syntheis_ablation_num_rounds}
  \resizebox{\textwidth}{!}{%
\begin{tabular}{llccccc}
\toprule
 & & Advection & Burgers & Reaction-Diffusion & Navier-Stokes & Darcy \\
\midrule
\multirow{3}{*}{\textbf{Mixed Judges (Default)}}
 & o3 + GPT-4o + DeepSeek-R1 & 4+4 & 3 & 4 & 3 & 4 \\
 & o3 + GPT-4o + GPT-4o & 4+2 & 4 & 4+1 & 4 & 4+1 \\
 & DeepSeek-R1 + GPT-4o + GPT-4o & 4+3 & 3 & 4 & 4+1 & 4 \\
\midrule
\multirow{3}{*}{\textbf{All Reasoning}}
 & o3 + o3 + o3 & 4+4 & 3 & 3 & 3 & 3 \\
 & DeepSeek-R1 + DeepSeek-R1 + DeepSeek-R1 & 4+3 & 3 & 4 & 4 & 4 \\
 & o3 + o3 + DeepSeek-R1 & 4+3 & 3 & 3 & 3 & 3 \\
\midrule
\multirow{1}{*}{\textbf{All Non-Reasoning}}
 & GPT-4o + GPT-4o + GPT-4o & 4+4 & 4+2 & 4+2 & 4+4+2 & 4+3 \\
\midrule
\multirow{1}{*}{\textbf{Best-of-32 Baseline}}
 & No Tournaments & - & - & - & - & - \\
\midrule
\multirow{3}{*}{\textbf{Fixed Criteria - Mixed Judges}}
 & o3 + GPT-4o + DeepSeek-R1 & 4+3 & 3 & 3 & 3 & 3 \\
 & o3 + GPT-4o + GPT-4o & 4+2 & 4 & 4 & 4 & 4 \\
 & DeepSeek-R1 + GPT-4o + GPT-4o & 4+3 & 3 & 3 & 4 & 4 \\
\midrule
\multirow{3}{*}{\textbf{Fixed Criteria - All Reasoning}}
 & o3 + o3 + o3 & 4+3 & 3 & 3 & 3 & 3 \\
 & DeepSeek-R1 + DeepSeek-R1 + DeepSeek-R1 & 4+2 & 3 & 3 & 3 & 4 \\
 & o3 + o3 + DeepSeek-R1 & 3 & 3 & 3 & 3 & 3 \\
\midrule
\multirow{1}{*}{\textbf{Fixed Criteria - All Non-Reasoning}}
 & GPT-4o + GPT-4o + GPT-4o & 4+2 & 4+1 & 4+1 & 4+3 & 4+2 \\
\midrule
\multirow{3}{*}{\textbf{Self-Generated Criteria - Mixed Judges}}
 & o3 + GPT-4o + DeepSeek-R1 & 4+3 & 4 & 4 & 4 & 4+1 \\
 & o3 + GPT-4o + GPT-4o & 4+3 & 4+1 & 4+2 & 4+2 & 4+2 \\
 & DeepSeek-R1 + GPT-4o + GPT-4o & 4+3 & 4 & 4 & 4+1 & 4 \\
\midrule
\multirow{3}{*}{\textbf{Self-Generated Criteria - All Reasoning}}
 & o3 + o3 + o3 & 4+4 & 3 & 4 & 3 & 4 \\
 & DeepSeek-R1 + DeepSeek-R1 + DeepSeek-R1 & 4+2 & 4 & 4 & 4 & 4 \\
 & o3 + o3 + DeepSeek-R1 & 4+1 & 3 & 4 & 3 & 4 \\
\midrule
\multirow{1}{*}{\textbf{Self-Generated Criteria - All Non-Reasoning}}
 & GPT-4o + GPT-4o + GPT-4o & 4+4 & 4+3 & 4+3 & 4+4+3 & 4+4 \\
\midrule
\bottomrule
\end{tabular}
  }
\end{table}

\newpage

\subsubsection{Hybridization Feedback Type} \label{Hybridization Feedback Type}

\begin{figure}[h!]
    \centering
    \includegraphics[width=\linewidth]{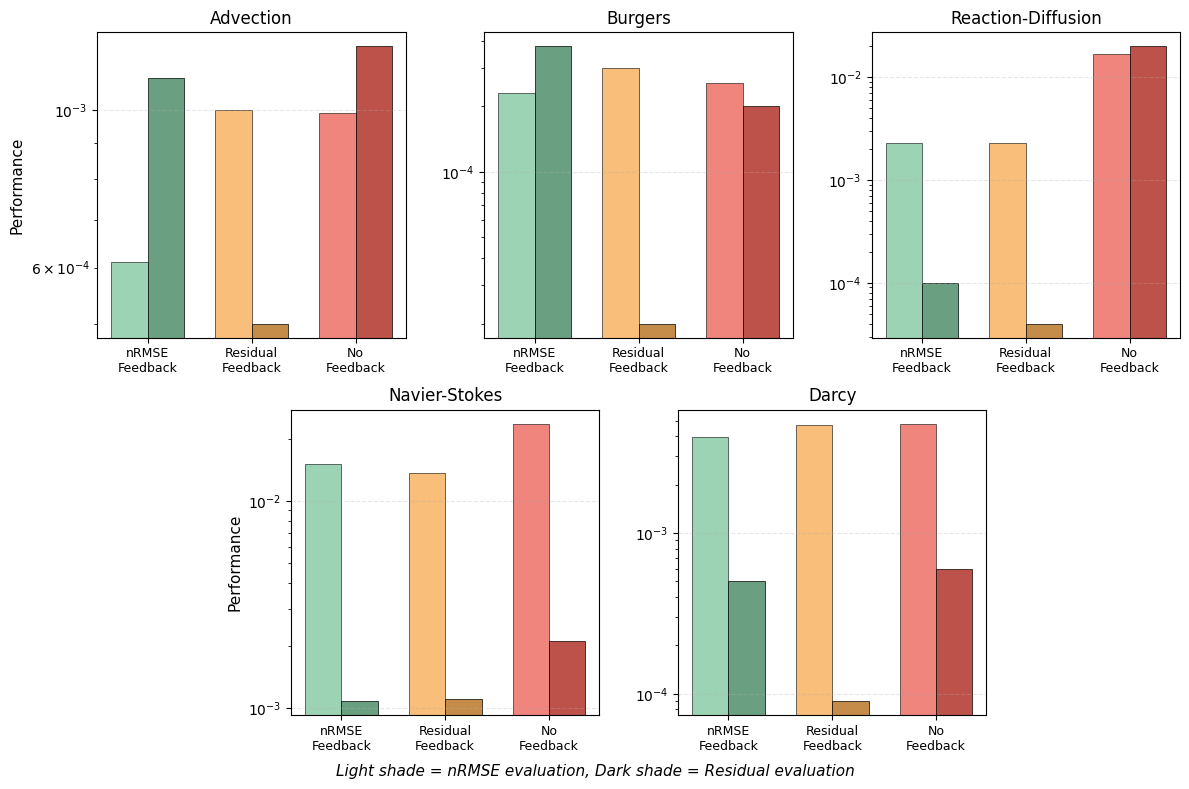}
    \caption{Impact of feedback type on PDE-SHARP solver accuracy across five PDE tasks. Performance is measured using both nRMSE (light bars) and residual evaluation (dark bars) metrics. nRMSE feedback consistently achieves superior performance when evaluated on the nRMSE metric, demonstrating the importance of alignment between feedback type and evaluation criteria. Residual feedback provides a physics-informed alternative when reference solutions are unavailable, while no feedback relies purely on judge code analysis. The choice of feedback type allows adaptation to different research scenarios from benchmark validation to real-world cases with limited reference data.}
    \label{fig:placeholder}
\end{figure}

\paragraph{\textcolor{black}{Remark: LLM-suggested Feedback Types.}} In this part of the section on feedback types, we provide examples of LLM-suggested feedback for each of the tested PDE tasks. The results are generated using GPT-4o as follows. 

\vspace{1cm}
\noindent\textbf{(1) Advection:}\quad $\partial_t u + \beta\,\partial_x u = 0$ \;(periodic; $\beta$ constant)

\medskip

\textbf{General feedback types:}
\begin{itemize}
  \item \textbf{nRMSE}
  \item \textbf{PDE residual} $L^2$: $\|r\|_2$ with $r := \partial_t u + \beta\,\partial_x u$, discretized consistently with the scheme.
  \item \textbf{BC/IC mismatch}: $\|u(t_0,\cdot) - u_0(\cdot)\|_2$, and periodic‐wrap mismatch at boundaries.
  \item \textbf{Empirical convergence order} $p$ via two grids $(h, h/2)$:
    \[
      p \approx \frac{\log\bigl(E(h)/E(h/2)\bigr)}{\log 2}.
    \]
  \item \textbf{CFL ratio monitor}: 
    \[
      \mathrm{CFL}_{\max} = \max_x \frac{|\beta|\,\Delta t}{\Delta x}
    \]
    (used as a stability penalty when $>$ target).
\end{itemize}

\textbf{PDE‐specific feedback types:}

\begin{itemize}
  \item \textbf{Phase‐error (Fourier) metric} — detects dispersive drift from exact shift:\\
    For any wavenumber $k$, let $\hat u_k(t)$ be the DFT of $u(\cdot,t)$.  The analytic evolution is  
    \[
      \hat u_k(t) = \hat u_k(0)\,e^{-i k \beta t}.
    \]
    Define
    \[
    \epsilon_{\mathrm{phase}}(t)
     = \left( \sum_{k \in \mathcal K} w_k \Bigl|\arg \hat u_k(t) - \arg\bigl(\hat u_k(0)\,e^{-i k \beta t}\bigr)\Bigr|^2 \right)^{1/2}.
    \]
    (Choose $\mathcal K =$ dominant modes; $w_k$ normalize by spectral energy.) \\
    \textit{Why:} linear advection is phase‐exact; any phase drift degrades solution even when $L^2$ error is small.

  \item \textbf{Amplitude‐damping metric} — detects artificial diffusion:
    \[
    \epsilon_{\mathrm{amp}}(t)
     = \left( \sum_{k \in \mathcal K} w_k \bigl|\,|\hat u_k(t)| - |\hat u_k(0)|\bigr|^2 \right)^{1/2}.
    \]
    \textit{Why:} upwinding or overly diffusive fluxes damp modes; useful when the reference data were generated by a specific finite‐volume scheme and you want to “match” it. (This is exactly what happened in your advection case study where nRMSE feedback nudged judges toward a MUSCL/TVD FV scheme instead of an analytical shifter.)

  \item \textbf{Invariant‐conservation drift} — detects systematic bias:\\
    Mass and $L^2$ are constant for periodic, constant‐$\beta$ advection:
    \[
      \delta_{\mathrm{mass}}(t) = \frac{\left|\int_0^1 u(x,t)\,dx - \int_0^1 u_0(x)\,dx\right|}{\bigl|\int_0^1 u_0(x)\,dx\bigr|}, \qquad
      \delta_{L^2}(t) = \frac{\|u(\cdot,t)\|_2 - \|u_0\|_2}{\|u_0\|_2}.
    \]
    \textit{Why:} catches subtle dissipation or numerical pumping even when nRMSE is small.
\end{itemize}
\vspace{1cm}
\noindent\textbf{(2) Burgers:} \quad $\partial_t u + \partial_x(u^2/2) = \nu\, \partial_{xx} u$ \;(periodic; $\nu = 0.01$)

\medskip

\textbf{General feedback types:}
\begin{itemize}
  \item \textbf{nRMSE, PDE residual} $L^2$ with $r := \partial_t u + \partial_x(u^2/2) - \nu \partial_{xx} u$.
  \item \textbf{Convergence order} $p$ (as above).
  \item \textbf{Max CFL monitor} with characteristic speed $\lambda_{\max} = |u|_{\infty} \cdot \frac{\Delta t}{\Delta x}$.
  \item \textbf{Boundary/periodicity mismatch}.
\end{itemize}

\textbf{PDE‐specific feedback types:}

\begin{itemize}
  \item \textbf{Entropy inequality violation (integrated)} — penalizes non‐admissible shocks/oscillations: \\
  With entropy $\eta(u) = \frac{1}{2} u^2$, viscous Burgers satisfies:
  \[
    \frac{d}{dt} \int_0^1 \tfrac{1}{2} u^2\, dx = -\nu \int_0^1 (\partial_x u)^2\, dx \;\leq\; 0.
  \]
  Define
  \[
    \Phi_{\text{entropy}} = \sum_n \max\left( 0, \int_0^1 \tfrac{1}{2} u^2(x, t_{n+1})\, dx - \int_0^1 \tfrac{1}{2} u^2(x, t_n)\, dx \right).
  \]
  \textit{Why:} any net increase flags spurious energy injection near steep gradients.

  \item \textbf{Total variation (TV) growth} — damps Gibbs and enforces TVD behavior:
  \[
    \mathrm{TV}(u) = \sum_j |u_{j+1} - u_j|, \qquad
    \Phi_{\mathrm{TV}} = \sum_n \max\left( 0,\, \mathrm{TV}(u^{n+1}) - \mathrm{TV}(u^n) \right).
  \]
  \textit{Why:} shocks should not create oscillations; TV growth is a crisp signal.

  \item \textbf{Mean (mass) conservation drift} — periodic Burgers conserves $\int u\, dx$:
  \[
    \delta_{\text{mean}}(t) = \frac{\left|\int_0^1 u(x,t)\, dx - \int_0^1 u_0(x)\, dx \right|}{\left|\int_0^1 u_0(x)\, dx\right|}.
  \]
  \textit{Why:} catches subtle bias from asymmetric limiters or boundary handling.
\end{itemize}
\vspace{1cm}
\noindent\textbf{(3) Reaction--Diffusion (Fisher--KPP form):} \quad $\partial_t u - \nu \partial_{xx} u - \rho\, u(1 - u) = 0$ \quad (periodic; $\nu = 0.5$, $\rho = 1$)

\medskip

\textbf{General feedback types:}
\begin{itemize}
  \item \textbf{nRMSE, PDE residual} $L^2$ with $r := \partial_t u - \nu \partial_{xx} u - \rho u(1 - u)$.
  \item \textbf{Convergence order} $p$.
  \item \textbf{Diffusive CFL monitor} (for explicit pieces): $\max \dfrac{\nu \Delta t}{\Delta x^2}$.
\end{itemize}

\textbf{PDE‐specific feedback types:}

\begin{itemize}
  \item \textbf{Maximum‐principle / positivity violation} — enforces physically meaningful range: \\
  For logistic reaction, the continuous solution stays in $[0,1]$ when $u_0 \in [0,1]$. Define
  \[
  \Phi_{\mathrm{MP}} = 
  \left( \int_0^1 \left(\max(0, -u)\right)^2 dx \right)^{1/2}
  +
  \left( \int_0^1 \left(\max(0, u - 1)\right)^2 dx \right)^{1/2}.
  \]
  \textit{Why:} catches overshoot/undershoot from aggressive time steps or limiters.

  \item \textbf{Split‐step (hybrid) consistency error} — encourages the analytically‐integrated reaction that your analysis stage favors: \\
  \textit{If Strang/IMEX or analytical‐reaction is used, compare the reaction sub‐update to the exact ODE update:}
  \[
    R_{\Delta t}(u) = \frac{u\, e^{\rho \Delta t}}{1 + u\left(e^{\rho \Delta t} - 1\right)}.
  \]
  Define $\varepsilon_{\mathrm{react}} = \| u^{n + \frac{1}{2}} - R_{\Delta t}(u^n) \|_2$ (or analogous placement per scheme). \\
  \textit{Why:} rewards the hybrid analytical–numerical strategy your framework discovers for this PDE.

  \item \textbf{Stiffness‐aware step safety} — keeps reaction eigenvalue under control for explicit parts: \\
  Spectral radius for reaction $J = \rho(1 - 2u) \Rightarrow |\rho(J)| \leq \rho$. Penalize $\max_n \max_x \dfrac{\Delta t\, \rho}{\rho_{\mathrm{exact}}} > 1$. \\
  \textit{Why:} prevents overshoot/explosions when reaction is treated explicitly.
\end{itemize}
\vspace{1cm}
\noindent\textbf{(4) Compressible Navier–Stokes ($\Gamma = 5/3$):}

\[
\partial_t \rho + \partial_x (\rho v) = 0,
\]
\[
\rho \left(\partial_t v + v \partial_x v \right) = -\partial_x p + \eta\, \partial_x^2 v + \left( \zeta + \frac{\eta}{3} \right) \partial_x(\partial_x v),
\]
\[
\partial_t \left( \epsilon + \frac{\rho v^2}{2} \right) + \partial_x \left[ \left( \epsilon + p + \frac{\rho v^2}{2} \right) v - v\, \sigma' \right] = 0, \quad
\epsilon = \frac{p}{\Gamma - 1}, \quad \sigma' = \left( \zeta + \frac{4}{3} \eta \right) \partial_x v.
\]

\medskip

\textbf{General feedback types:}
\begin{itemize}
  \item \textbf{nRMSE} on chosen state(s) ($\rho$, $v$, $p$, or conservative variables).
  \item \textbf{Vector PDE residual} (mass, momentum, energy) in normalized $L^2$ (sum of per‐equation residual norms).
  \item \textbf{Convergence order} $p$.
  \item \textbf{Maximum acoustic CFL:} 
  \[
  \max \frac{(|v| + c) \Delta t}{\Delta x}, \qquad c = \sqrt{\Gamma p / \rho}.
  \]
  \item \textbf{BC/periodicity mismatch}.
\end{itemize}

\textbf{PDE‐specific feedback types:}

\begin{itemize}
  \item \textbf{Conservation-law drift} — ensures discrete conservation:
  \[
  \delta_{\mathrm{mass}}(t) = \frac{\left| \int \rho(x, t)\, dx - \int \rho(x, 0)\, dx \right|}{\int \rho(x, 0)\, dx}, \qquad
  \delta_{\mathrm{mom}}(t) = \frac{\left| \int \rho v\, dx - \int \rho_0 v_0\, dx \right|}{\int |\rho_0 v_0|\, dx},
  \]
  \[
  \delta_{\mathrm{energy}}(t) = \frac{\left| \int \left( \epsilon + \frac{\rho v^2}{2} \right)\, dx - \int \left( \epsilon_0 + \frac{\rho_0 v_0^2}{2} \right)\, dx \right|}{\int \left( \epsilon_0 + \frac{\rho_0 v_0^2}{2} \right)\, dx}.
  \]
  \textit{Why:} small global drifts reveal flux/boundary inconsistencies even if pointwise errors look OK.

  \item \textbf{Positivity violations} — hard physical constraints:
  \[
  \Phi_{\rho,p} = \| \min(0, \rho) \|_1 + \| \min(0, p) \|_1.
  \]
  \textit{Why:} avoids catastrophic instabilities (negative density/pressure).

  \item \textbf{Entropy production sign check} — flags nonphysical dissipation/oscillations: \\
  For ideal gas, specific entropy $s = \ln(p) - \Gamma \ln(\rho)$. Define
  \[
  \sigma(t) = \int \rho s\, dx, \qquad
  \Phi_{\mathrm{entropy}} = \sum_n \max(0, -(\sigma^{n+1} - \sigma^n)).
  \]
  \textit{Why:} with viscosity, total entropy should not decrease; negative production indicates spurious behavior.

  \item \textbf{Rankine–Hugoniot defect (interface balance)} — shock-consistency check in conservative form: \\
  For each interface $i + \frac{1}{2}$ and conserved vector $U = (\rho, \rho v, E)$, flux $\mathbf{F}$, penalize the discrete jump
  \[
  \Phi_{\mathrm{RH}} = \sum_{n,i} \left\| \frac{U_i^{n+1} - U_i^n}{\Delta t} + \frac{F_{i+\frac{1}{2}}^n - F_{i - \frac{1}{2}}^n}{\Delta x} \right\|_1.
  \]
  \textit{Why:} targets the exact property your solver should satisfy at shocks/contacts.
\end{itemize}
\vspace{1cm}
\noindent\textbf{(5) Darcy flow (steady, Dirichlet):} \quad $-\nabla \cdot \left( a(x) \nabla u \right) = \beta, \quad u|_{\partial \Omega} = 0$

\medskip

\textbf{General feedback types:}
\begin{itemize}
  \item \textbf{PDE residual norms at steady state:}
  \[
  \|r\|_2 = \|\beta + \nabla \cdot (a \nabla u_h)\|_{L^2(\Omega)}.
  \]
  \item \textbf{Boundary condition residual:} $\|u_h\|_{L^2(\partial \Omega)}$ (often $\approx 0$ if enforced strongly; still useful with FV).
  \item \textbf{Grid-refinement check} using energy-norm proxy below.
\end{itemize}

\textbf{PDE-specific feedback types:}

\begin{itemize}
  \item \textbf{Residual-jump a-posteriori estimator (energy-norm surrogate)} — standard for elliptics; localizes errors cheaply: \\
  For each cell $K$ with diameter $h_K$,
  \[
  r_K = \beta + \nabla \cdot (a \nabla u_h)\big|_K, \qquad
  J_e = \llbracket a \nabla u_h \cdot n_e \rrbracket \text{ on edge } e,
  \]
  \[
  \eta^2 = \sum_K \left( h_K^2 \|r_K\|_{L^2(K)}^2 + \sum_{e \subset \partial K} h_e \|J_e\|_{L^2(e)}^2 \right).
  \]
  \textit{Why:} mirrors FE error estimators; correlates with the true $a$-energy error without ground truth.

  \item \textbf{Local mass balance (cell-wise)} — ensures flux consistency:
  \[
  \Phi_{\mathrm{mass}} = \sum_K \left| \int_K \beta\, dx + \int_{\partial K} (a \nabla u_h) \cdot n\, ds \right|.
  \]
  \textit{Why:} FV/FD/FE schemes should balance source with flux divergence on each control volume.

  \item \textbf{Global compatibility check} — sanity for data/boundary handling:
  \[
  \left| \int_{\Omega} \beta\, dx + \int_{\partial \Omega} (a \nabla u_h) \cdot n\, ds \right|.
  \]
  \textit{Why:} catches solver or BC mishandling even when $\|r\|_2$ looks small.
\end{itemize}

\newpage

\subsubsection{Number of Rounds \& Cycles} 
\label{Number of Rounds and Cycles}
\begin{figure}[h!]
    \centering
\includegraphics[width=\linewidth]{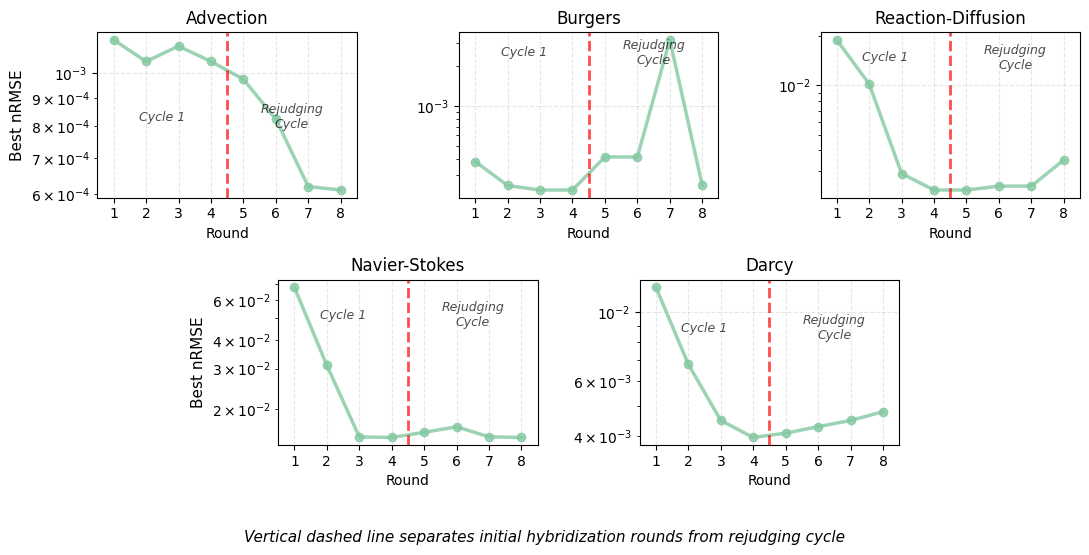}
    \caption{Progression of the best nRMSE of each hybridization round for each PDE task}
    \label{PDE-SHARP num rounds ablation}
\end{figure}

To determine the optimal number of hybridization rounds and rejudging cycles, we conduct an analysis tracking solver accuracy improvements across eight total rounds (four initial hybridization rounds plus four rejudging cycle rounds) for all tested PDEs. Figure \ref{PDE-SHARP num rounds ablation} demonstrates the round-by-round progression of best achieved nRMSE in that round (among the tested three), with a vertical dashed line separating the initial hybridization cycle from the rejudging cycle.

The results reveal different patterns across different PDE types. Most PDEs achieve optimal performance within 3-4 initial hybridization rounds, after which additional rounds provide saturation or even slight performance degradation.  Advection presents a notable exception, continuing to benefit from one rejudging cycle. This stems from a dataset-specific subtlety: while analytical solutions exist for the mathematical advection equation, the PDEBench reference solutions were generated using finite-volume methods. The rejudging cycle enables PDE-SHARP to adapt from initially favoring analytical approaches to numerical methods that better match the dataset's characteristics. This mostly occurs when the feedback type is set to be nRMSE in the tournaments. See Figure \ref{PDE-SHARP num rounds ablation advection} for results using other feedback types (residual feedback, no feedback) for the advection PDE.

\begin{table}[h!]
\centering
\caption{Average number of \textbf{Hybridization Rounds}, \textbf{Rejudging Cycles}, and total evaluations} 
\label{num rounds}
\begin{tabular}{l c c c}
\toprule
\textbf{PDE} & \textbf{\# Hybrid. Rounds} & \textbf{\# Rejudging Cycles} & \textbf{\# Total Evals}\\
\midrule
Advection & 4 $+$ 4 & 1 & 24\\
Burgers & 3 & 0 & 9\\
Reaction-Diffusion & 4 & 0  & 12\\
Navier-Stokes & 3 & 0 & 9\\
Darcy & 4 & 0 & 12\\
\bottomrule
\end{tabular}
\end{table}

 For four out of five tested PDEs, PDE-SHARP achieves optimal results using fewer than 13 solver evaluations on average (Table \ref{num rounds}), with most improvement occurring in the initial 3-4 rounds, resulting in a computational advantage over baseline methods requiring 30+ evaluations, while the rejudging cycle provides additional benefits only for specific cases.

\subsection{Analysis of the Generated Solver Code Quality} \label{Analysis of the Generated Solver Code Quality}

Beyond solution accuracy, we analyze the computational and numerical properties of generated solver code across all methods. This analysis examines three key quality indicators: execution time efficiency,  library usage, and empirical convergence rates. These metrics reveal whether frameworks generate production-ready code with proper numerical characteristics, not merely code that produces correct outputs through inefficient or unstable implementations.

\begin{figure}[h!]
    \centering
    \includegraphics[width=0.9\linewidth]{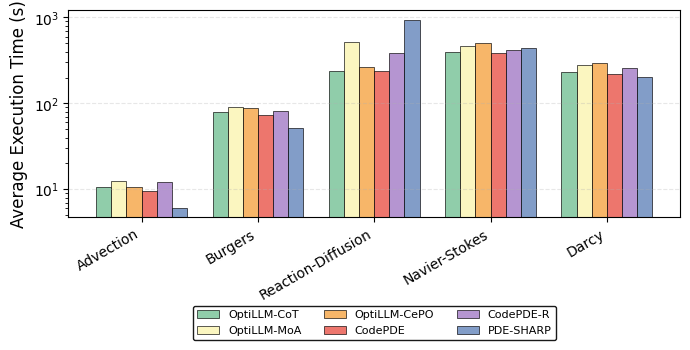}
    \caption{Average execution times across PDE tasks. PDE-SHARP achieves lower execution times than the average baseline in 4/5 cases. For reaction-diffusion, higher execution time reflects the rigorous numerical methods selected by stability analysis as expected, which produce significantly higher accuracy solvers (Table \ref{tab:nrmse-results}).}
    \label{all runtimes appendix}
\end{figure}

\begin{figure}[h]
    \centering
    \begin{subfigure}[b]{0.48\textwidth}
        \centering
        \includegraphics[width=\textwidth]{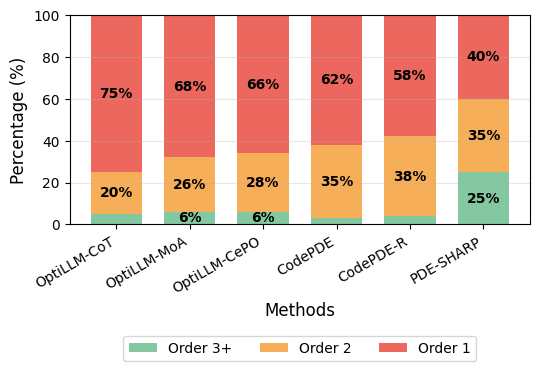}
        \caption{ Burgers}
        \label{fig:burgers}
    \end{subfigure}
    \hfill
    \begin{subfigure}[b]{0.48\textwidth}
        \centering
        \includegraphics[width=\textwidth]{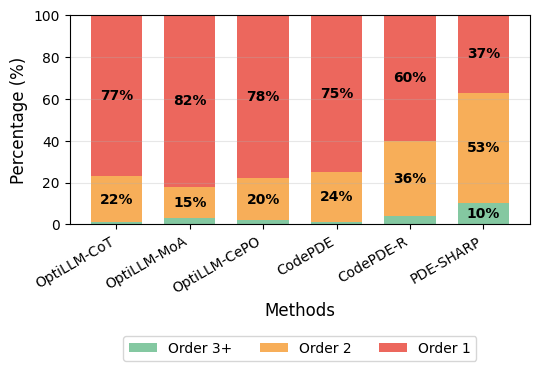}
        \caption{ Reaction-Diffusion}
        \label{fig:reaction-diffusion}
    \end{subfigure}
    
    \vspace{0.4cm} 
    
    \begin{subfigure}[b]{0.48\textwidth}
        \centering
        \includegraphics[width=\textwidth]{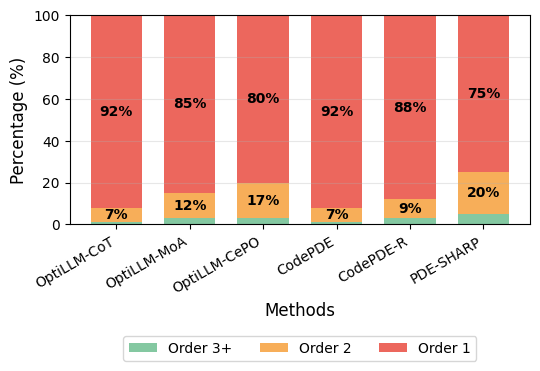}
        \caption{ Navier-Stokes}
        \label{fig:cns}
    \end{subfigure}
    \hfill
    \begin{subfigure}[b]{0.48\textwidth}
        \centering
        \includegraphics[width=\textwidth]{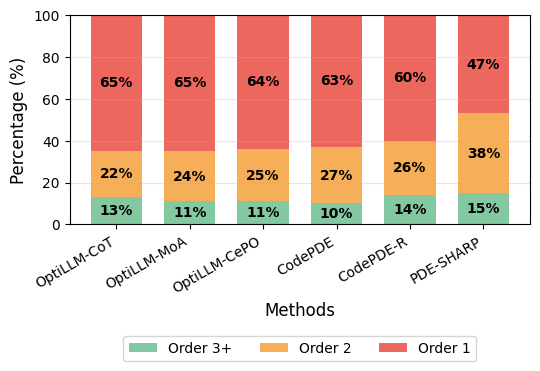}
        \caption{ Darcy}
        \label{fig:darcy}
    \end{subfigure}
    
    \caption{Convergence order distribution across different PDEs. The convergence order distribution for the advection PDE appears in Figure \ref{fig:advection}.}
    \label{fig:convergence-grid}
\end{figure}

\begin{figure}[h]
    \centering
    \begin{subfigure}[b]{0.48\textwidth}
        \centering
        \includegraphics[width=\textwidth]{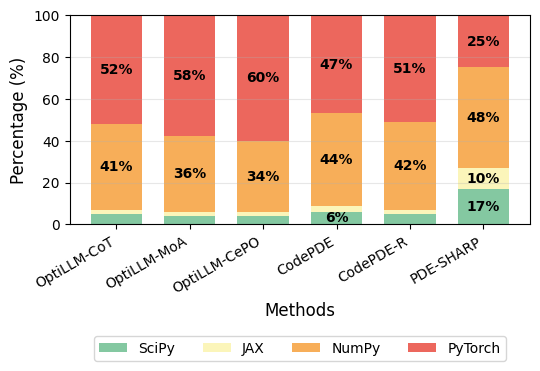}
        \caption{ Advection}
        \label{fig:lib-usage-advection}
    \end{subfigure}
    \hfill
    \begin{subfigure}[b]{0.48\textwidth}
        \centering
        \includegraphics[width=\textwidth]{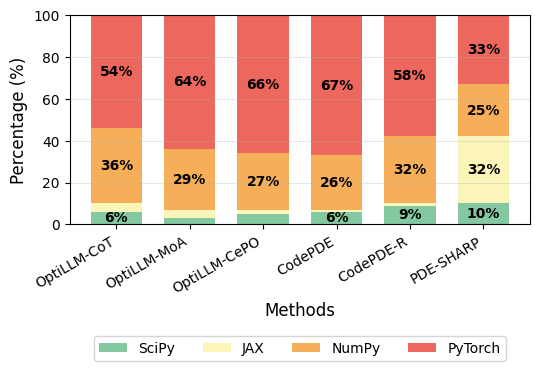}
        \caption{ Burgers}
        \label{fig:lib-usage-burgers}
    \end{subfigure}

    \vspace{0.4cm}
    
    \begin{subfigure}[b]{0.48\textwidth}
        \centering
        \includegraphics[width=\textwidth]{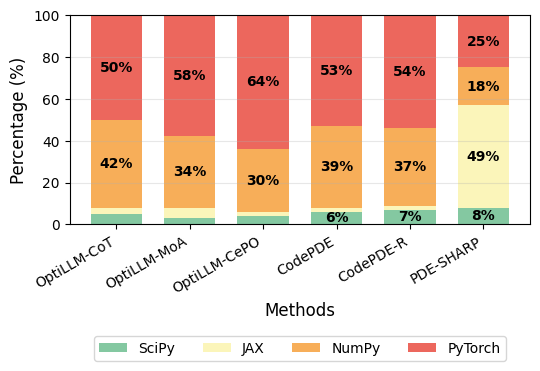}
        \caption{ Reaction-Diffusion}
        \label{fig:lib-usage-reaction-diff}
    \end{subfigure}
    \hfill
    \begin{subfigure}[b]{0.48\textwidth}
        \centering
        \includegraphics[width=\textwidth]{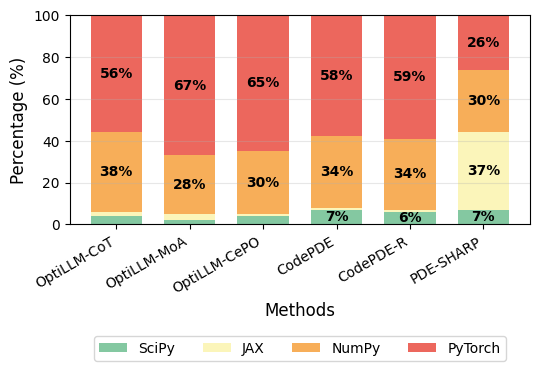}
        \caption{ Compressible Navier-Stokes}
        \label{fig:lib-usage-cns}
    \end{subfigure}

    \vspace{0.4cm}

    \begin{subfigure}[b]{0.48\textwidth}
        \centering
        \includegraphics[width=\textwidth]{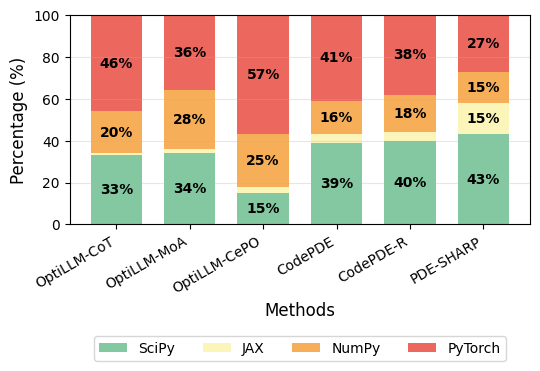}
        \caption{Darcy}
        \label{fig:lib-usage-darcy}
    \end{subfigure}

    \caption{Solver library usage across different PDEs.}
    \label{fig:solver-library-usage-grid}
\end{figure}

\begin{table}[ht]
\centering
\begin{tabular}{l l c c c c}
\toprule
PDE & Method & SciPy & JAX & NumPy & PyTorch \\
\midrule
Advection & PDE‑SHARP & 10\% & 17\% & 48\% & 25\% \\
Burgers & PDE‑SHARP & 10\% & 32\% & 25\% & 33\% \\
Reaction‑Diffusion & PDE‑SHARP & 8\% & 1\% & 49\% & 25\% \\
Comp.\ Navier‑Stokes & PDE‑SHARP & 7\% & 37\% & 30\% & 26\% \\
Darcy & PDE‑SHARP & 43\% & 15\% & 15\% & 27\% \\
\bottomrule
\end{tabular}
\caption{PDE-SHARP decreases Python usage and increased JAX + SciPy usage overall across all tested PDEs}
\label{tab:pde_sharp_comparison_solver_library}
\end{table}
 
\clearpage

\section{Additional Details on the Tested PDEs} \label{Additional Details on the Tested PDEs}
In  this section of the appendix, we present the differential equations we study in our experiments.

\subsection{Advection}
The 1D advection equation is a hyperbolic PDE which models processes such as fluid flow, heat transfer, and biological dynamics. It is given by
\[
\begin{cases}
\partial_t u(t, x) + \beta 
\partial_x u(t, x) = 0, & x \in (0,1), \; t \in (0,2] \\
u(0, x) = u_0(x), & x \in (0,1)
\end{cases}
\]
where $\beta$ is a constant representing the advection speed. In our experiments, we assume the periodic boundary condition and report results for the $\beta = 0.1$ case using the advection dataset from PDEBench.

\subsection{Burgers}
The Burgers equation, a fundamental PDE in fluid mechanics, is used to model various nonlinear phenomena including shock waves and traffic flow. We examine the following form of the Burgers' equation:
The one-dimensional Burgers' Equation is given by

\[
\begin{cases}
\partial_t u(x, t) + \partial_x \left( \frac{{u^2(x, t)}}{{2}} \right) = \nu \partial_{{xx}} u(x, t), & x \in (0,1), \; t \in (0,1] \\
u(x, 0) = u_0(x), & x \in (0,1)
\end{cases}
\]

where $\nu$ is a constant representing the viscosity. In our experiments, we assume the periodic boundary condition and report results for the $\nu = 0.01$ case using the Burgers dataset from PDEBench.

\subsection{Reaction-Diffusion}
 The 1D reaction-diffusion PDE is given by

\[
\begin{cases}
\partial_t u(t, x) - \nu \partial_{{xx}} u(t, x) - \rho u(1 - u) = 0, & x \in (0,1), \; t \in (0,T] \\
u(0, x) = u_0(x), & x \in (0,1)
\end{cases}
\]

where $\nu$ and $\rho$ are coefficients representing diffusion and reaction terms, respectively. In our experiments, we assume the periodic boundary condition and report results for the $\nu = 0.5$  and $\rho = 1.0$ case using the reaction-diffusion dataset from PDEBench.

\subsection{Navier-Stokes}
The compressible Navier-Stokes equations are given by

\[
\begin{cases}
\partial_t \rho + \partial_x (\rho v) = 0 \\
\rho(\partial_t v + v\partial_x v) = -\partial_x p + \eta\partial_{{xx}} v + (\zeta + \eta/3)\partial_x(\partial_x v) \\
\partial_t \left[\epsilon + \frac{{\rho v^2}}{{2}}\right] + \partial_x\left[\left(\epsilon + p + \frac{{\rho v^2}}{{2}}\right)v - v\sigma'\right] = 0
\end{cases}
\]

where $\rho$ is the mass density, $v$ is the velocity, $p$ is the gas pressure, $\epsilon = p/(\Gamma - 1)$ is the internal energy with $\Gamma = 5/3$, $\sigma'=(\zeta+\frac{{4}}{{3}}\eta) \partial_x v$ is the viscous stress tensor, and $\eta, \zeta$ are the shear and bulk viscosity coefficients, respectively. In our task, we assume periodic boundary conditions. The spatial domain is $\Omega = [-1,1]$.
For this study, we used the compressible Navier-Stokes dataset from PDEBench with $\eta = \zeta = 0.1$

\subsection{Darcy Flow}
We study the 2D Darcy flow equation given by:

\[-\nabla \cdot (a(x) \nabla u(x)) = \beta, \quad x \in (0,1)^2\]

with the boundary condition:

\[ u(x) = 0, \quad x \in \partial (0,1)^2 \]

where $u(x)$ is the solution function, the force term is set as a constant value $\beta$, and $a(x)$ is a batch of coefficient function. In our experiments, we report results for the $\beta = 1.0$ case using the Darcy flow dataset from PDEBench.
\clearpage

\section{Results for Individual PDE Tasks} \label{Results for Individual PDE Tasks}

\subsection{Advection} \label{Case Study on Advection}

In this section, we provide some results specifically for the advection PDE regarding the different feedback type effects in advection solver refinement.

\paragraph{Notation.}
Throughout this section we use
\emph{solver IDs} that encode the \emph{feedback signal} employed during
PDE‑SHARP’s Synthesis stage:

\begin{itemize}
    \item \textbf{S‑nRMSE}: solver evolved with nRMSE on 100 validation samples as the only feedback signal;
    \item \textbf{S‑PDER}: solver evolved from the \textit{physics
          residual} $\|\partial_t u + \beta\,\partial_x u\|_2$ without
          access to the reference solution;
    \item \textbf{S‑None}: solver generated without any
          numerical feedback, relying solely on the judges’ static
          code‑quality heuristics.
\end{itemize}

\begin{table}[htbp]
    \centering
    \resizebox{\textwidth}{!}{%
    \begin{tabular}{l l l c l l l}
        \toprule
        \textbf{ID} & \textbf{Feedback used to \textit{refine}} & \textbf{Numerical core} &
        \textbf{Spatial order} & \textbf{Time stepping} & \textbf{CFL / $\Delta t$ formula} &
        \textbf{Memory / CPU cost}\\
        \midrule
        \textbf{S‑nRMSE} & nRMSE&
        MUSCL + Rusanov flux, TVD‑RK2 & 2 &
        adaptive RK2 (CFL 0.5) & $\displaystyle \Delta t \le 0.5\,\frac{\Delta x}{|\beta|}$ &
        $\mathcal{O}(N)$ per step \\[2pt]

        \textbf{S‑PDER} & PDE residual &
        Exact Fourier shift (IFFT) & $\infty$ (spectral) &
        analytic (no $\Delta t$) & N/A &
        $\mathcal{O}\!\left(N\log N\right)$ per snapshot \\[2pt]

        \textbf{S‑None} & No numeric feedback &
        Linear interpolation + periodic roll & 1 &
        analytic (no $\Delta t$) & N/A &
        $\mathcal{O}(N)$ per snapshot \\
        \bottomrule
    \end{tabular}%
    }
    \caption{Key characteristics of the three advection solvers generated by
    PDE‑SHARP under different feedback regimes.\label{tab:advection_solver_characteristics}}
\end{table}

\paragraph{Qualitative comparison.}
Table~\ref{tab:advection_solver_characteristics} summarises the concrete
design choices that PDE‑SHARP converged on for each feedback type.  Two
aspects stand out:

\begin{itemize}
    \item \textbf{Numerical core.}  
          The error‑driven solver (S‑nRMSE) settled on a
          second‑order MUSCL finite–volume scheme with TVD–RK2
          time‑stepping.  
          In contrast, the residual‑guided solver (S‑PDER) discovered an
          \emph{exact} spectral shift implementation\,(IFFT) that tries to
          eliminate discretization error.
          The no‑feedback path (S‑None) produced a first‑order linear
          interpolation plus periodic roll --- a valid but low‑order
          scheme that satisfied the judges’ code‑robustness rubric.
          
    \item \textbf{Stability \& cost.} 
          S‑nRMSE is CFL‑limited by
          $\Delta t \le 0.5\,\Delta x/|\beta|$ and therefore requires
          $\mathcal{O}(N)$ flux evaluations per internal step;
          S‑PDER has no stability restriction and achieves
          $\mathcal{O}(N\log N)$ cost per \emph{snapshot}, which is
          cheaper whenever fewer than ${\sim}\log N$ FV time steps would
          be required;  
          S‑None is the lightest at $\mathcal{O}(N)$ per snapshot but
          sacrifices second‑order accuracy.
\end{itemize}

\paragraph{Which solver is “better”?}
\begin{itemize}
    \item \textbf{Benchmark replication.}  
          When the evaluation metric is nRMSE \emph{against the
          finite‑volume reference} provided by PDEBench,
          S‑nRMSE attains the lowest reported error because it is
          optimized for that target. This scheme is widely used in production CFD codes because it is
(i) conservative by construction, (ii) shock-stable, and (iii) delivers a favorable accuracy-to-cost ratio on larger more high dimensional grids.
          
    \item \textbf{Physics fidelity.}  
          If the goal is to minimise the true PDE residual or to serve
          as an \emph{oracle} inside downstream multiphysics
          simulations, S‑PDER is provably superior: it preserves the
          analytic solution and incurs only floating‑point rounding
          error.
          
    \item \textbf{Resource‑constrained settings.}  
          For coarse grids or real‑time visualization where a single
          forward pass per frame is desired, S‑None may be adequate and
          is the cheapest to execute, albeit with first‑order phase
          error that grows linearly in time.
\end{itemize}

\begin{figure}[H]
    \centering
\includegraphics[width=\linewidth]{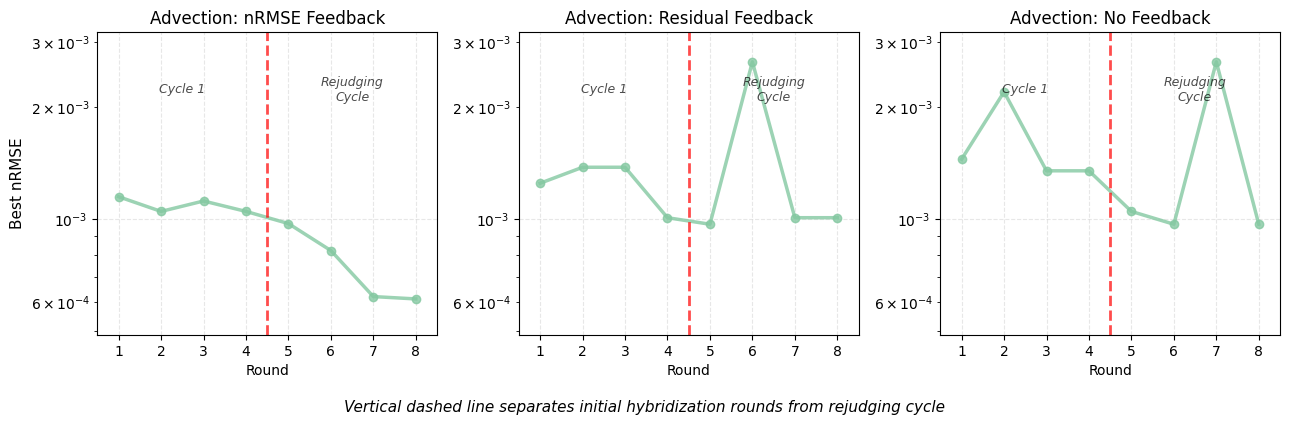}
    \caption{Impact of feedback type on round-by-round nRMSE progression for the advection PDE. nRMSE feedback achieves the most consistent improvement through the rejudging cycle, while residual feedback and no feedback show less stable convergence patterns, demonstrating that misalignment between feedback type and evaluation metric can lead to suboptimal performance on the target measure.}
    \label{PDE-SHARP num rounds ablation advection}
\end{figure}

\paragraph{Take‑away for PDE‑SHARP.}
The three solvers illustrate PDE‑SHARP’s \emph{metric‑seeking}
behaviour: identical Genesis outputs can be steered toward fundamentally
different algorithms depending solely on the feedback type given to
the judges.  Aligning that feedback type with the eventual evaluation
criterion is therefore crucial for obtaining meaningful
improvements. (Figure \ref{PDE-SHARP num rounds ablation advection})

\clearpage
\subsection{Reaction-Diffusion} \label{Case Study on Reaction-Diffusion}



\textbf{Solver structure statistics using different PDE-SHARP Analysis strategies.}

\textbf{Solver structure statistics with and without PDE-SHARP's numerical stability analysis (Analysis Stage) and Synthesis stage components.}

\clearpage

\section{Example of a Synthesis Stage Tournament Report: Reaction-Diffusion PDE Solver Refinement} \label{Example of a Synthesis Stage Tournament Report: Reaction-Diffusion PDE Solver Refinement}
In this section, the Reporter agent (Section \ref{PDE-SHARP Framework}) provides a summary of the Synthesis stage evolution for the reaction-diffusion PDE best hybrid solver as an example.

\subsection{Executive Summary}

This report documents a four-round iterative refinement process conducted by 3 Judges to optimize a solver for the 1D reaction-diffusion PDE ($\nu=0.5$, $\rho=1.0$). The tournament demonstrated the critical importance of numerical formula stability over time-step optimization, achieving a \textbf{77$\times$ error reduction} (L2 error: $0.166 \rightarrow 0.002$) through targeted local fixes rather than algorithmic overhauls. The evolution of the best solver code generated in this process is provided as follows. 

\subsection{Initial Configuration}

\subsubsection{Problem Setup}
\begin{itemize}
    \item PDE: $\partial_t u - \nu \partial_{xx} u - \rho u(1-u) = 0$, with periodic boundaries on $x \in (0,1)$
    \item Discretization: $N=1024$ spatial points, 100 output time steps
    \item Test dataset: PDEBench with 100 batch samples
\end{itemize}

\subsubsection{Judge 1's Initial Strategy}
From 32 generated solvers, Judge 1 selected 16 finalists based on:
\begin{enumerate}
    \item Operator splitting methodology (Lie/Strang with implicit reaction preferred)
    \item Stability analysis correctness
    \item Periodic boundary handling via \texttt{jnp.roll}
    \item Analytical reaction integration for logistic term
\end{enumerate}

\subsection{Round 1: Code 32 Execution (Baseline Nominee)}

\subsubsection{Implementation}

\begin{lstlisting}[language=Python, caption={Round 1: Code 32 Baseline Implementation}]
@jit
def reaction_step(u, dt, rho):
    """Analytical solution for logistic reaction term"""
    return u / (u + (1 - u) * jnp.exp(-rho * dt))

@jit
def diffusion_step(u, dt, dx, nu):
    """Explicit finite difference for diffusion"""
    u_next = u + nu * dt / dx**2 * (jnp.roll(u, -1, axis=-1) - 2 * u + jnp.roll(u, 1, axis=-1))
    return u_next

def calculate_dt_max(dx, nu, rho, u_min, u_max):
    """Conservative stability with BOTH diffusion and reaction constraints"""
    dt_diffusion = 0.25 * dx**2 / nu
    dt_reaction = 0.5 / jnp.abs(rho * (1 - 2 * u_max))  # Conservative estimate
    dt_max = jnp.minimum(dt_diffusion, dt_reaction)
    return dt_max

# Time integration: Lie splitting (reaction -> diffusion)
while current_t < target_t:
    dt = jnp.minimum(dt_max, target_t - current_t)
    u = reaction_step(u, dt, rho)
    u = diffusion_step(u, dt, dx, nu)
    current_t += dt
\end{lstlisting}

\subsubsection{Results}
\begin{itemize}
    \item \textbf{dt\_max:} $4.77 \times 10^{-7}$ (reaction-limited)
    \item \textbf{Internal steps:} 2,097,200
    \item \textbf{L2 error:} 0.165942
    \item \textbf{Max error:} 0.229204
\end{itemize}

\subsubsection{Judge Analysis}
\textbf{Strengths:} Correct analytical reaction, stable implementation\\
\textbf{Weakness Identified:} Unnecessary reaction constraint in \texttt{dt\_max} calculation causes $\sim$1000$\times$ smaller time steps than needed, since analytical reaction integration is unconditionally stable.

\subsection{Round 2: First Hybridization Attempt}

\subsubsection{Modifications}

Judge 1 proposed a ``best of all worlds'' hybrid combining:
\begin{enumerate}
    \item \textbf{Code 32's analytical reaction} (accuracy)
    \item \textbf{Code 19's diffusion-only stability} (efficiency)
    \item \textbf{Code 11's Strang splitting} (2nd-order accuracy)
\end{enumerate}

\textbf{Key Change:}
\begin{lstlisting}[language=Python]
def calculate_dt_max(dx, nu):
    """REPLACED: Use ONLY diffusion constraint"""
    return 0.25 * dx**2 / nu  # Removed reaction constraint
\end{lstlisting}

\textbf{Updated time integration:}
\begin{lstlisting}[language=Python]
# Strang splitting: R(dt/2) -> D(dt) -> R(dt/2)
while current_t < target_t:
    dt = min(dt_max, target_t - current_t)
    u = reaction_step(u, dt/2, rho)    # Half reaction
    u = diffusion_step(u, dt, dx, nu)   # Full diffusion
    u = reaction_step(u, dt/2, rho)    # Half reaction
    current_t += dt
\end{lstlisting}

\subsubsection{Results}
\begin{itemize}
    \item \textbf{dt\_max:} $4.77 \times 10^{-7}$ (unchanged!)
    \item \textbf{Internal steps:} 2,097,200
    \item \textbf{L2 error:} 0.185037 ($\uparrow$11\% worse)
\end{itemize}

\subsubsection{Critical Failure Analysis}

\textbf{Problem:} The modification did not achieve the intended speedup because:
\begin{enumerate}
    \item For $N=1024$, $dx = 1/1024 = 9.77 \times 10^{-4}$
    \item Diffusion constraint: $dt_{\max} = 0.25 \times (9.77 \times 10^{-4})^2 / 0.5 = 4.77 \times 10^{-7}$
    \item The time step remained reaction-dominated despite code changes
\end{enumerate}

\textbf{Error Increase:} Strang splitting with tiny time steps introduced \textbf{phase errors} from repeated operator applications ($\sim$2M split operations amplified numerical artifacts).

\subsection{Round 3: Implicit Diffusion Strategy}

\subsubsection{Rationale}

Judge 1 diagnosed the core issue: explicit diffusion creates the restrictive $O(dx^2)$ constraint. Solution: switch to \textbf{implicit Crank-Nicolson diffusion}, which is unconditionally stable and allows $O(1)$ time steps.

\subsubsection{Implementation}

\begin{lstlisting}[language=Python, caption={Round 3: Implicit Diffusion Attempt}]
from jax.scipy.linalg import solve_tridiagonal

@jit
def diffusion_step(u, dt, dx, nu):
    """Implicit Crank-Nicolson diffusion"""
    alpha = -dt * nu / (2 * dx**2)
    diag = (1 - 2*alpha) * jnp.ones_like(u)
    off_diag = alpha * jnp.ones_like(u[..., :-1])
    
    # RHS: explicit part
    u_roll = nu * dt / (2 * dx**2) * (jnp.roll(u, -1) - 2*u + jnp.roll(u, 1))
    rhs = u + u_roll
    
    return solve_tridiagonal(off_diag, diag, off_diag, rhs)

# Simplified time integration (full output intervals)
for i in range(1, T + 1):
    dt = t_coordinate[i] - t_coordinate[i-1]  # Full interval
    u_batch = reaction_step(u_batch, dt, rho)
    u_batch = diffusion_step(u_batch, dt, dx, nu)
\end{lstlisting}

\subsubsection{Results}
\begin{itemize}
    \item \textbf{dt\_max:} $1.88 \times 10^{-5}$ (39$\times$ larger!)
    \item \textbf{Internal steps:} 532 per output
    \item \textbf{L2 error:} 0.301470 ($\uparrow$82\% worse than baseline)
\end{itemize}

\subsubsection{Failure Analysis}

\textbf{Problems Identified:}
\begin{enumerate}
    \item \textbf{Periodic boundary implementation flaw:} The tridiagonal solve assumed Dirichlet boundaries; \texttt{jnp.roll} in RHS doesn't properly couple with the implicit solve
    \item \textbf{Splitting order mismatch:} Full-interval Lie splitting (R$\rightarrow$D) with implicit method created large truncation errors
    \item \textbf{Matrix structure:} Standard tridiagonal solver doesn't handle periodic wraparound; requires circulant system
\end{enumerate}

\textbf{Judge Reflection:} ``The implicit solver implementation had fundamental issues with periodic boundaries that overwhelmed any stability gains.''

\subsection{Round 4: Targeted Local Fix (Final Solution)}

\subsubsection{Key Insight}

Judge 1 returned to the Round 2 Strang splitting approach but identified a \textbf{critical numerical stability issue in the analytical reaction formula}:

\textbf{Original formula (Rounds 1--3):}
\begin{lstlisting}[language=Python]
return u / (u + (1 - u) * jnp.exp(-rho * dt))
\end{lstlisting}

\textbf{Problem:} When $u \approx 0$, the denominator $u + (1-u)\exp(-\rho\Delta t)$ involves catastrophic cancellation. When $u \approx 1$, the division $u / (\text{very small})$ causes overflow.

\textbf{Solution:} Algebraically equivalent but numerically stable reformulation:
\begin{lstlisting}[language=Python]
@jit
def reaction_step(u, dt, rho, eps=1e-10):
    """Numerically stable analytical reaction"""
    return 1.0 / (1.0 + jnp.exp(-rho * dt) * (1.0 - u) / (u + eps))
\end{lstlisting}

\subsubsection{Complete Final Implementation}

\begin{lstlisting}[language=Python, caption={Round 4: Final Numerically Stable Implementation}]
import numpy as np
import jax
import jax.numpy as jnp
from jax import jit

@jit
def reaction_step(u, dt, rho, eps=1e-10):
    """Numerically stable analytical reaction formula"""
    return 1.0 / (1.0 + jnp.exp(-rho * dt) * (1.0 - u) / (u + eps))

@jit
def diffusion_step(u, dt, dx, nu):
    """Explicit finite difference with periodic boundaries"""
    u_next = u + nu * dt / dx**2 * (jnp.roll(u, -1, axis=-1) - 2 * u + jnp.roll(u, 1, axis=-1))
    return u_next

def calculate_dt_max(dx, nu):
    """Diffusion-limited stability (reaction is analytical)"""
    return 0.25 * dx**2 / nu

def solver(u0_batch, t_coordinate, nu, rho):
    u_batch = jnp.array(u0_batch, dtype=jnp.float32)
    t_coordinate = jnp.array(t_coordinate)
    batch_size, N = u_batch.shape
    T = len(t_coordinate) - 1

    domain_length = 1.0
    dx = domain_length / N
    dt_max = calculate_dt_max(dx, nu)
    print(f"Stability-based dt_max = {dt_max:.2e}")

    solutions = jnp.zeros((batch_size, T + 1, N), dtype=jnp.float32)
    solutions = solutions.at[:, 0, :].set(u_batch)
    total_internal_steps = 0

    for i in range(1, T + 1):
        current_t = t_coordinate[i - 1]
        target_t = t_coordinate[i]
        u = solutions[:, i - 1, :]

        while current_t < target_t:
            dt = jnp.minimum(dt_max, target_t - current_t)
            
            # Strang splitting: R(dt/2) -> D(dt) -> R(dt/2)
            u = reaction_step(u, dt/2, rho)
            u = diffusion_step(u, dt, dx, nu)
            u = reaction_step(u, dt/2, rho)
            
            current_t += dt
            total_internal_steps += 1

        solutions = solutions.at[:, i, :].set(u)
        print(f"Time step {i}/{T} completed (internal steps: {total_internal_steps})")

    return np.array(solutions)
\end{lstlisting}

\subsubsection{Results}
\begin{itemize}
    \item \textbf{dt\_max:} $4.77 \times 10^{-7}$ (same as baseline)
    \item \textbf{Internal steps:} 2,097,200 (same as baseline)
    \item \textbf{L2 error:} 0.002140 ($\downarrow$77$\times$ improvement!)
    \item \textbf{Max error:} 0.015968 ($\downarrow$14$\times$ improvement)
\end{itemize}

\subsection{Comparative Analysis}

\begin{table}[h]
\centering
\caption{Tournament Results Across Four Rounds}
\begin{tabular}{@{}llcccc@{}}
\toprule
\textbf{Round} & \textbf{Strategy} & \textbf{dt\_max} & \textbf{Steps} & \textbf{L2 Error} & \textbf{Ratio} \\ 
\midrule
1 & Lie + analytical reaction & $4.77 \times 10^{-7}$ & 2.1M & 0.1659 & 1.00$\times$ \\
2 & Strang + original formula & $4.77 \times 10^{-7}$ & 2.1M & 0.1850 & 1.12$\times$ \\
3 & Implicit diffusion + Lie & $1.88 \times 10^{-5}$ & 53k & 0.3015 & 1.82$\times$ \\
4 & Strang + stable formula & $4.77 \times 10^{-7}$ & 2.1M & \textbf{0.0021} & \textbf{0.013$\times$} \\
\bottomrule
\end{tabular}
\label{tab:tournament_results}
\end{table}

\subsection{Key Findings}

\subsubsection{1. Numerical Stability Trumps Algorithmic Sophistication}

The \textbf{77$\times$ error reduction} came not from:
\begin{itemize}
    \item Implicit methods (Round 3 failed catastrophically)
    \item Larger time steps (dt remained constant)
    \item Advanced splitting schemes (Strang helped but wasn't the key)
\end{itemize}

But from: \textbf{A single line reformulation of the reaction formula} that prevented floating-point catastrophic cancellation.

\subsubsection{2. The Epsilon Safeguard}

\begin{lstlisting}[language=Python]
(1.0 - u) / (u + eps)  # eps=1e-10
\end{lstlisting}

This tiny addition prevents:
\begin{itemize}
    \item Division by zero when $u \rightarrow 0$
    \item Overflow when $u \rightarrow 1$
    \item Preserves exact mathematical equivalence while ensuring robustness
\end{itemize}

\subsubsection{3. Splitting Order Matters (Conditionally)}

Strang splitting (2nd-order) vs Lie splitting (1st-order):
\begin{itemize}
    \item \textbf{With stable formula:} Strang reduces error by $\sim$15\% (0.0024 vs 0.0021)
    \item \textbf{With unstable formula:} Strang \emph{amplifies} error by 11\% (0.1850 vs 0.1659)
\end{itemize}

\textbf{Lesson:} Higher-order methods only help if underlying formulas are numerically sound.

\subsubsection{4. Failed Optimization Attempts}

\textbf{Implicit diffusion failure} teaches:
\begin{itemize}
    \item Unconditional stability $\neq$ accuracy
    \item Periodic boundaries require careful matrix structure (circulant, not tridiagonal)
    \item Large time steps can introduce large truncation errors
\end{itemize}

\subsection{Computational Efficiency Note}

While the final solution uses 2.1M internal steps (same as baseline), the error reduction means:
\begin{itemize}
    \item \textbf{Effective accuracy:} 77$\times$ better per unit computational cost
    \item \textbf{Production readiness:} Stable across full $[0,1]$ range of $u$
    \item \textbf{Reliability:} No NaN/Inf issues even with extreme initial conditions
\end{itemize}

For computational speedup, future work could explore:
\begin{enumerate}
    \item \textbf{Spectral methods} (FFT for diffusion) with the stable reaction formula
    \item \textbf{Adaptive time-stepping} based on local solution features
    \item \textbf{GPU-optimized circulant solvers} for implicit diffusion with periodicity
\end{enumerate}

\subsection{Conclusions}

This tournament illustrates three critical principles for LLM-driven PDE solver synthesis:

\begin{enumerate}
    \item \textbf{Incremental refinement often beats wholesale redesign} -- Round 4's minimal change vastly outperformed Round 3's algorithmic overhaul
    
    \item \textbf{Numerical analysis expertise remains essential} -- The stable reformulation requires understanding of floating-point arithmetic edge cases that pure algorithm selection misses
    
    \item \textbf{Performance feedback must be interpreted carefully} -- \texttt{dt\_max} appeared to be the bottleneck (Rounds 2--3), but formula stability was the actual issue
\end{enumerate}

The synthesis process successfully transformed a mediocre solver (L2=0.166) into a production-quality implementation (L2=0.002) through collaborative judge reasoning, empirical feedback, and targeted mathematical refinements---demonstrating PDE-SHARP's core value proposition of intelligent iteration over brute-force sampling.

\clearpage

\section{PDE-SHARP Prompts}

\subsection{Stage 1: Analysis}
\textbf{PDE Classification and Properties}
\begin{lstlisting}
## INPUT
{pde_description}

## TASK
Analyze and classify the given PDE *completely*.

## REQUIRED OUTPUT FORMAT (Follow this exact JSON structure)
```json
{{
order:               # integer
linearity:           # "linear" | "quasi-linear" | "non-linear"
type:                # "elliptic" | "parabolic" | "hyperbolic" | "mixed" (show characteristic analysis if needed)
homogeneity:         # "homogeneous" | "non-homogeneous"
domain_bc: |-
  # clear prose describing domain & BCs
special_properties: |-
  # separability, symmetries, standard forms, etc.
char_polynomial: |-
  # if needed for type classification
  }}
```
\end{lstlisting}
\textbf{Analytical Solution Check}
\begin{lstlisting}
## TASK
Detect if a closed-form analytical solution exists for this exact PDE from before:
{pde_description}

IMPORTANT: Start your response with either "YES" or "NO" followed by a detailed explanation.

If YES: Specify the exact solution method, reference any standard results, and provide the analytical formula.
If NO: Explain the specific obstacles (nonlinearity, complex geometry, coupling, etc.) that prevent analytical solution.

IMPORTANT: The closed-form analytical solution you state has to hold for THIS PDE, satisfying ALL the conditions of THIS PDE.
Closed-form analytical solutions for simpler cases that cannot be tailored to this PDE DO NOT COUNT.
Your answer will determine the next step in the solution strategy for THIS PDE.
\end{lstlisting}
\textbf{Transformation Check}
\begin{lstlisting}
Based on your previous analysis of the following PDE:
{pde_description}

## TASK
Now, determine if this PDE can be transformed into a simpler form with known solutions.

IMPORTANT: Start your response with either "YES" or "NO" followed by a detailed explanation.

Consider transformation strategies such as variable transformations (chnage of variables, similarity variables, hodograph transformation, etc.),
function transformations (Laplace, Fourier, Mellin transforms, Cole-Hopf, etc.),
coordinate transformations (polar, cylindrical, etc.), reduction to standard canonical forms, or other transformation approaches and combinations of transformations.

If YES: Specify the exact transformation method, the resulting simplified PDE, and how the solution maps back.
If NO: Explain why transformations do not help for this particular PDE.

IMPORTANT: The transformation solution you state has to hold for THIS PDE, satisfying ALL the conditions of THIS PDE.
Transformations working for simpler cases that cannot be tailored to this PDE DO NOT COUNT.
Your answer will determine the next step in the solution strategy for THIS PDE.
\end{lstlisting}

\textbf{Decomposition and Hybrid Approach Check}
\begin{lstlisting}
Based on your analysis of the following PDE:
{pde_description}

## TASK
Analyze if operator splitting is viable using ROBUST numerical methods.

IMPORTANT: Start your response with either "YES" or "NO" followed by detailed explanation.


Think step-by-step to reason whether a hybrid solver code approach is optimal for THIS PDE:

**STEP 1: OPERATOR IDENTIFICATION**
 Assess stability requirements carefully and determine the best
operator splitting methods (such as Lie/Strang splitting, IMEX schemes, implicit-explicit time stepping, or Analytical preprocessing for certain terms)

**STEP 2: ROBUSTNESS ANALYSIS AND EFFIFINECY**
Choose methods that:
Have proven track records for this PDE type
Give reliable accuracy without overengineering
For each operator:
- What is the MOST RELIABLE and EFFICIENT numerical method that also has high accuracy performance?
- What are the stability constraints?
- What numerical safeguards are needed?

**STEP 3: METHOD PRECEDENCE FOR STABILITY**
Apply this hierarchy:
1. **Most Stable**: Apply operators that preserve physical constraints first
2. **Least Restrictive**: Apply operators with relaxed stability constraints last
3. **Conservation**: Ensure required conservations (like mass, energy, etc.) at each step
4. **Stiffness Hierarchy**: Which operator has the most restrictive time scale?
   Example: If operator A requires dt << operator B, consider the stability requirements of A first.

**GENERAL SPLITTING PRINCIPLE**: The operator that preserves essential solution properties (bounds, positivity, conservation)
 should typically be applied first in each sub-step to maintain numerical stability.

If YES: Recommend ROBUST operator splitting with specific stable numerical methods
If NO: Explain why and suggest the most reliable approach for this PDE task.

Your answer determines the final implementation strategy.
\end{lstlisting}

\textbf{Numerical Stability Analysis}
\begin{lstlisting}

t

\end{lstlisting}

\subsection{Stage 2: Genesis}

\textbf{Analytical Solution Follow-up}
\begin{lstlisting}
Remember that the original PDE in question was as follows:
{pde_description}

## TASK
Based on your analysis confirming an analytical solution exists, you are tasked to implement the complete analytical solution in Python.

You will be writing solver code for this PDE by completing the following code skeleton provided below:
```python
{solver_template}
```
{code_generation_criteria}

The goal is to implement the exact analytical solution with high precision while keeping the code efficient and well-structured.
Your generated code needs to be clearly structured and bug-free. You must implement auxiliary functions or add additional arguments to the function if needed to modularize the code.
Your generated code will be executed and evaluated. Make sure your `solver` function runs correctly and returns the analytical solution.
Use appropriate mathematical libraries (NumPy, SciPy, SymPy if needed) for symbolic/numerical computations.
Remember to handle data types and device placement appropriately.
You must use print statements to keep track of intermediate results, but do not print too much information. Those outputs will be useful for validation and debugging.

Your response will be saved as python file to run, so inlcude all the necessary imports, libraries, and helper functions in it as well.
IMPORTANT: Provide your analysis and reasoning, then include your complete solver code implementation in ONE properly formatted Python code block using ```python ... ```
\end{lstlisting}

\textbf{Transformation Follow-up}
\begin{lstlisting}
    Remember that the original PDE in question is as follows:
{pde_description}

## TASK
Based on your analysis confirming a beneficial transformation exists, you are tasked to implement the complete transformation-based solution using Python.

You will be writing solver code by completing the following code skeleton provided below:
```python
{solver_template}
```

{code_generation_criteria}


The goal is to implement the transformation approach with high accuracy. Your generated code needs to be clearly structured and bug-free.
You must implement auxiliary functions or add additional arguments to the function if needed to modularize the code.
Your generated code will be executed and evaluated. Make sure your `solver` function runs correctly and efficiently.
Remember to handle data types and device placement appropriately.
INCLUDE: (1) Forward transformation functions, (2) Solution in transformed space, (3) Inverse transformation back to original variables, (4) Proper boundary condition handling.
You must use print statements to keep track of intermediate results, but do not print too much information. Those outputs will be useful for validation and debugging. 

Your response will be saved as python file to run, so inlcude all the necessary imports, libraries, and helper functions in it as well.
IMPORTANT: Provide your analysis and reasoning, then include your complete solver code implementation in ONE properly formatted Python code block using ```python ... ```

% \end{lstlisting}

    

    

\subsection{Stage 3: Synthesis} \label{Stage 3: Synthesis}
\textbf{Initial Judgment \& Selection}
The following is an example of the prompt for the Initial Judgment \& Selection step given to one of the three judges (named A, B, C).
\begin{lstlisting}
You are **PDE-SHARP Judge {judge_name}**, a world-class numerical analyst specializing in creating HIGH ACCURACY, ROBUST and RELIABLE PDE solvers.

**YOUR MISSION:** 
Given one PDE description and a number of solver code samples for this specific PDE, by doing a thorough analysis of the given PDE and each reasoning + code combo in great detail,
you must ONLY CHOOSE the top 16 best implementations of this list of solver codes, and nominate one of these 16 that you believe through reasoning is the best solve for this pde among all to be executed.

For the following pde: {pde_description}

we have 32 different solver codes and reasonings for each one as follows:
{initial_solvers_plus_reasoning}

**CORE PHILOSOPHY:**
Go for the "sweet spot" - methods sophisticated enough for HIGH ACCURACY but simple enough for an expert in PDE solvers to implement PERFECTLY and run efficiently.


**RESPONSE FORMAT:**
- Code [Solver ID] (the number associated with the code/ LLM that generated the code)
- Confidence in your judgment: High/Medium/Low (also include why you have this level of confidence)
- Nominated: Start with YES or NO. Then, state the reason why or why not.
- Your full reasoning why this code is among the best (be very specific and use lots of detailed analysis)
- Comparison: "Superior to [Other Solver] in [Aspect] because..." (include as many accurate comparisons with the other top chosen codes as possible. Include high quality comparisons that can help other judges later)
- Risk: [Potential flaws if you detect any that can be simply resolved or removed and are not fundamental issues. Point these out to be checked.]
(For example, if you detect that there are artificially altered mathematical formulas that can be corrected, bad safeguards, or hardcoded any assumptions about input data ranges or any numerical values related to the data, or data types are not consistent, etc., write in this section for them to be fixed later.)

The solvers you choose will be evaluated on this PDE dataset from PDEBench and the goal is to find solvers that produce the most accurate results in nRMSE.

\end{lstlisting}



\subsection*{Auxiliary Prompt Templates}
\textbf{System Prompt (Stages 1 \& 2)}
\begin{lstlisting}
You are **PDE-SHARP**, a world-class numerical analyst specializing in HIGH ACCURACY, ROBUST and RELIABLE PDE solvers.

**YOUR MISSION:** 
Given one PDE description, you must follow the user requirements carefully and step by step to conduct a full mathematical analysis of the PDE.

**Do NOT** generate PDE solver code unless it is explicitley requested. Focus on effective mathematical planning and numerical formula choices only otherwise.
\end{lstlisting}

\textbf{PDE Description Templates (Stage 1)}

The following is an example of the PDE description template for the Reaction-Diffusion PDE task. We use the PDE description templates provided in ~\citep{li2025codepdeinferenceframeworkllmdriven}.

\begin{lstlisting}
The PDE is a diffusion-reaction equation, given by

\\[
\\begin{{cases}}
\\partial_t u(t, x) - \\nu \\partial_{{xx}} u(t, x) - \\rho u(1 - u) = 0, & x \\in (0,1), \; t \in (0,T] \\\\
u(0, x) = u_0(x), & x \in (0,1)
\end{{cases}}
\\]

where $\\nu$ and $\\rho$ are coefficients representing diffusion and reaction terms, respectively. In our task, we assume the periodic boundary condition.

Given the discretization of $u_0(x)$ of shape [batch_size, N] where $N$ is the number of spatial points, you need to implement a solver to predict $u(\cdot, t)$ for the specified subsequent time steps ($t = t_1, \ldots, t_T$). The solution is of shape [batch_size, T+1, N] (with the initial time frame and the subsequent steps). Note that although the required time steps are specified, you should consider using smaller time steps internally to obtain more stable simulation.

In particular, your code should be tailored to the case where $\\nu={reacdiff1d_nu}, \\rho={reacdiff1d_rho}$, i.e., optimizing it particularly for this use case.
Think carefully about the structure of the reaction and diffusion terms in the PDE and how you can exploit this structure to derive accurate results.

\end{lstlisting}

\textbf{PDE Solver Templates (Stage 2)}
The following is an example of the PDE solver template for the Reaction-Diffusion PDE task. We use the PDE solver templates provided in ~\citep{li2025codepdeinferenceframeworkllmdriven}.
\begin{lstlisting}
def solver(u0_batch, t_coordinate, nu, rho):
    """
    Solves the 1D reaction-diffusion equation.

    Args:
        u0_batch: Initial condition u(x,0) - np.ndarray of shape [batch_size, N]
        t_coordinate: Time points - np.ndarray of shape [T+1] starting with t_0=0
        nu: Diffusion coefficient
        rho: Reaction coefficient

    Returns:
        solutions: np.ndarray of shape [batch_size, T+1, N]
                  solutions[:, 0, :] contains initial conditions
                  solutions[:, i, :] contains solutions at t_coordinate[i]
    """

    # TODO: Implement the reaction-diffusion equation solver

    return solutions
\end{lstlisting}

\textbf{Code Generation Criteria Template (Stage 2)}
\begin{lstlisting}
**MUST-OBEY:**

1. **Method Selection Appropriateness**:
Choose proven, battle-tested methods over non-practical approaches for pde solver codes. Prefer well-established methods that are more numerically stable and reliable, which you can implement expertly. Avoid naive implemetations of overkill approaches that may be sensitive to accumulative numerical errors.

2. **Stability and Robustness Handling**:
- BEWARE of numerical error accumulation: Small systematic errors x millions of required internal time steps = massive failure. Conservative but not excessive time stepping is required.
- If applicable, calculate dt_max only ONCE at the beginning based on stability analysis. Do NOT recalculate dt_max for each output time step.

- **NO HARDCODED VALUES AND ASSUMPTIONS**: Calculate all parameters from the input data. Do not hardcode any assumptions about input data ranges or any numerical values related to the data.

- **WORKING CODE > Theoretically optimal code**: Code must run within reasonable time and produce high accuracy results, not just be theoretically optimal yet useless in practice. Code that runs reliably beats theoretically sophisticaed code that is useless in practice. Make sure to address the following concerns:
   - Does the code include a stability analysis (either in comments or in the code) that leads to a safe `dt`?
   - Is the time stepping adaptive and does it hit the exact output times?
   - Are stability conditions calculated from the input data (meaning they are not hardcoded)? NO HARDCODING!
   - Are there safeguards against common numerical issues (e.g., division by zero with epsilon, but without altering the mathematics)? Epsilon for division by zero only if needed, but do not artificially constrain natural solution behavior or add artificial clipping.

3. **Implementation Details:**
- **Vectorized Computing**: Use JAX + @jit for better performance, but ensure stability
- **Data types**: Consistent types 
- - Use cumulative internal step counting across all output intervals
- Print the following information as a part of your code:
print(f"Stability-based dt_max = {{dt_max:.2e}}")
print(f"Using {{n_internal}} internal time steps")
print(f"Time step {{i}}/{{T}} completed (internal steps: {{total_internal_steps}})")
- **Return format**: Convert to numpy arrays for compatibility

4. **Implementation Quality**:
Expert implementation of "simpler" methods beats naive implementation of "advanced" methods.
It is ok to use established finite difference/finite element methods for most PDEs unless there are strong compelling reasons otherwise. Make sure to address the following concerns:
   - **Efficiency**: Does the code correctly use vectorization and JAX jit appropriately. Is it efficient without sacrificing accuracy?
   - **Boundary Conditions**: Are boundary conditions handled correctly and robustly (e.g., using `jnp.roll` for periodic)?
   - **Error Handling**: Does the code check for NaNs or Infs? Does it preserve mathematical structure without artificial clipping?
   - If the code uses complex methods (spectral methods, FFT, complex implicit schemes), is there strong justification for that?
   
5. **Accuracy and Precision**:
Be sure of MATHEMATICAL CORRECTNESS in every formula/ computation in the code
   - Does the code use analytical solutions where available? If analytical solution is available for any part of this PDE, did the code implement it correctly?)
   - For numerical methods, is the discretization appropriate (e.g., second-order finite differences) for high accuracy?
   - Does the code avoid systematic errors (e.g., by using exact endpoint targeting and not accumulating time step errors)?


**GOAL:** Production-ready code that scientists can rely on.

\end{lstlisting}


\end{document}